\documentclass[10pt, a4paper,conference]{IEEEtran}
\ifCLASSINFOpdf
\else
\fi
\usepackage[table]{xcolor} 
\usepackage{graphicx}
\usepackage{caption}
\usepackage{booktabs}
\usepackage{hyperref}
\usepackage{lipsum}
\usepackage{subfigure}
\usepackage{soul}
\usepackage{magicabbrev}
\usepackage{siunitx}
\usepackage{cancel}
\usepackage{amssymb}
\usepackage{multirow}
\DeclareSIUnit\px{px}
\usepackage{glossaries}
\newacronym[]{gls:RS}{RS}{remote sensing}
\newacronym[]{gls:GSD}{GSD}{ground sampling distance}
\newacronym[]{gls:CNN}{CNN}{convolutional neural network}
\newacronym[]{gls:CRF}{CRF}{conditional random field}
\newacronym[]{gls:SGD}{SGD}{Stochastic Gradient Descent}
\newacronym[]{gls:FCNN}{FCNN}{fully convolutional neural network}
\newacronym[]{gls:DCNN}{DCNN}{deep convolutional neural network}
\newacronym[]{gls:ACNN}{ACNN}{atrous convolutional neural network}
\newacronym[longplural={oriented bounding boxes}]{gls:OBB}{OBB}{oriented bounding box}
\newacronym[longplural={rotated bounding boxes}]{gls:RBB}{RBB}{rotated bounding box}
\newacronym[longplural={horizontal bounding boxes}]{gls:HBB}{HBB}{horizontal bounding box}
\newacronym[]{gls:RCNN}{RCNN}{Region-based convolutional neural network}

\newacronym[]{gls:miou}{mIoU}{mean intersection over union}

\newacronym[]{AV}{AV}{autonomous vehicles}
\newacronym[]{HD}{HD}{high definition}
\newacronym[]{ADAS}{ADAS}{advanced vehicle assistance system}
\newacronym[]{GPS}{GPS}{global positioning system}
\newacronym[]{USGS}{USGS}{U.S. Geological Survey}
\newacronym[]{DWT}{DWT}{discrete wavelet transform}
\newacronym[]{SVM}{SVM}{support vector machine}
\newacronym[]{DSM}{DSM}{digital surface model}
\newacronym[]{IU}{IU}{intersection over union}

\newglossaryentry{gls:FastRCNN}{name={Fast-\gls{gls:RCNN}}, description={}}
\newglossaryentry{gls:FasterRCNN}{name={Faster-\gls{gls:RCNN}}, description={}}
\newglossaryentry{gls:MaskRCNN}{name={Mask-\gls{gls:RCNN}}, description={}}
\newacronym[]{EAGLE}{EAGLE}{oriEnted object detection using Aerial imaGery in real-worLd scEnarios}
\newacronym[]{gls:UAV}{UAV}{unmanned aerial vehicles}
\newacronym[]{gls:iou}{IoU}{intersection over union}
\newacronym[]{gls:map}{mAP}{mean average precision}
\newacronym[]{gls:ap}{AP}{average precision}
\newacronym[]{gls:ar}{AR}{average recall}
\newacronym[]{gls:GFLOP}{GFLOP}{giga floating point operation}
\newacronym[]{gls:RPN}{RPN}{region proposal network}
\newacronym[]{gls:ROI}{ROI}{region of interest}
\newacronym[]{gls:FPN}{FPN}{feature pyramid network}
\newacronym[]{gls:HOG}{HOG}{histogram of oriented gradients}
\newacronym[]{gls:DAB}{DAB}{domain adapter block}
\newacronym[]{gls:NMS}{NMS}{non-maximum suppression}
\newacronym[]{gls:FPS}{FPS}{frame per second}
\newacronym[]{gls:SSD}{SSD}{single shot detector}
\newacronym[]{gls:ICN}{ICN}{image cascade network}
\newacronym[]{gls:DIN}{DIN}{deformable inception network}
\newacronym[]{gls:R-RPN}{R-RPN}{multi-scale rotational region-proposal network}
\newacronym[]{gls:R-ROI}{R-ROI}{multi-scale rotational region of interest network}
\newacronym[]{gls:R-NMS}{R-NMS}{rotational non-maximum suppression}
\newacronym[]{gls:OHEM}{OHEM}{online hard example mining}
\glsdisablehyper
\defglsentryfmt{
	\ifglsused{\glslabel}{
		\glsgenentryfmt
	}{
	\emph{\glsgenentryfmt}}}
\newcommand{\headImagesO}{
\makeatletter
\let\@oldmaketitle\@maketitle% Store \@maketitle
\renewcommand{\@maketitle}{\@oldmaketitle% Update \@maketitle to insert...
  \includegraphics[width=\linewidth,height=9\baselineskip]
    {figures/sample_val_long.pdf}\bigskip}% ... an image
\makeatother
}

\newcommand{\ResultsHBBandRBB}{%
\begin{table*}[t]
	\centering
	\vspace{0.2cm}
 	\caption{Benchmark of the state of the art on the horizontal bounding box (HBB) and the rotated bounding box (RBB) detection task; mAP means mean Average Precision, higher is better. Mask-RCNN-H means trained on horizontal bounding box. Mask-RCNN-R means trained on rotated bounding box.}
    \resizebox{0.9\textwidth}{!}{
	\begin{tabular}{cc|ccc|ccc}
		Method                                  & Backbone          & \multicolumn{3}{c}{AP~$[\%]$ (HBB)}  & \multicolumn{3}{c}{AP~$[\%]$ (RBB)}\\
		                                        &                   & Mean  & SV    & LV            & Mean    & SV   & LV\\% \Bstrut\\
	    \hline
        Yolov3~\cite{redmon2018yolov3}          & Darknet-53        & 20.29 & 30.45 & 10.13         & 13.28 & 21.34 &  5.23\\%\Tstrut\\
        SSD~\cite{liu2016ssd}                   & InceptionV2       & 12.06 & 20.67 &  3.45         &  7.31 & 12.34 &  2.28\\%\Tstrut\\
        RefineDet~\cite{zhang2018single}        & VGG16             & 22.23 & 32.25 & 12.21         & 14.78 & 22.67 &  6.89\\%\Tstrut\\ 
        R-FCN~\cite{dai2016r}                   & ResNet101         & 30.61 & 46.85 & 14.37         & 21.06 & 35.56 &  6.56\\%\Tstrut\\
        Faster-RCNN~\cite{fasterrcnnNIPS2015}   & ResNet101         & 31.84 & 48.34 & 15.34         & 23.15 & 39.29 &  7.02\\%\Tstrut\\  
        Mask-RCNN~\cite{he2017maskrcnn}         & ResNet101         & 30.81 & 46.51 & 15.11         & 22.54 & 36.65 &  8.43\\%\Tstrut\\   
        Cascade-RCNN~\cite{cai2018cascade}      & ResNet101         & 33.49 & 49.65 & 17.34         & 23.58 & 38.97 &  8.19\\%\Tstrut\\
        SNIPER~\cite{singh2018sniper}           & ResNet101         & 30.74 & 48.34 & 13.14         & 21.97 & 38.23 &  5.72\\%\Tstrut\\
        FPN~\cite{fpn}                          & ResNet101         & 37.10 & 50.76 & 23.45         & 27.11 & 39.78 & 14.45\\%\Tstrut\\
        TridenNet~\cite{li2019scale}            & ResNet101         & 30.53 & 47.16 & 13.91         & 22.53 & 37.16 &  7.91\\%\Tstrut\\
        FCOS~\cite{tian2019fcos}                & ResNeXt101        & 38.80 & 52.94 & 24.67         & 27.67 & 41.24 & 14.10\\%\Tstrut\\
        Cascade Mask-RCNN-H~\cite{cai2019cascade} & Triple-ResNeXt152 & \textbf{39.29} & \textbf{53.45} & \textbf{25.14}         & 30.22 & 43.84 & 16.60\\%\Tstrut\\
        \hline
        Cascade Mask-RCNN-R [Ours]              & Triple-ResNeXt152 & -     & -     & -             & \textbf{37.23} & \textbf{51.27} & \textbf{23.19}%\Tstrut\\
	\end{tabular}
	}\label{tab:ResultsHbbandRBB}
\end{table*}
}

\newcommand{\Resultssecondsplit}{%
\begin{table}[t]
	\centering
		 \caption{Benchmark of the best method from benchmark on the second split approach by splitting based on flight campgain. Cascae Mask-RCNN-O, -H, and -R means Cascade Mask-RCNN trained on oriented, horizontal, and rotated bounding boxes respectively.}
		\resizebox{0.49\textwidth}{!}{
		\begin{tabular}{c|c|c|c|c}
		\multicolumn{1}{c}{Method} & 
		\multicolumn{1}{c}{Task} & 
		\multicolumn{1}{c}{mAP $[\%]$} &
		\multicolumn{2}{c}{AP  $[\%]$}\\
		\multicolumn{1}{c}{} & 
        \multicolumn{1}{c}{} & 
		\multicolumn{1}{c}{} & 
        \multicolumn{1}{c}{small-vehicle} & 
        \multicolumn{1}{c}{large-vehicle}\\%\Bstrut\\
	 \hline
        Cascade Mask-RCNN-H & HBB  &  33.54 & 50.16 & 16.92\\
        Cascade Mask-RCNN-R & RBB & 30.18 & 46.82 & 13.54\\
        Cascade Mask-RCNN-O & OBB & 32.02 & 48.13 & 15.91
	\end{tabular}
	\vspace{-0.8cm}
	}\label{tab:Resultssecondsplit} 
\end{table}
}

\newcommand{\Resultscrossvalidation}{\begin{table}
\footnotesize
\centering
\caption{Comparison of results on EAGLE and DOTA using Cascade Mask-RCNN.The comparison is based on mAP. SL and LV stand for small-vehicle and large-vehicles respectively. (scores in mAP)}
\label{tab:generalization}
\begin{tabular}{c|c||c|c|c}
Training set & Test set & Avg. & SV & LV\\
\hline
DOTA    & DOTA  & 59.95 & 61.23 & 58.67\\
DOTA    & EAGLE & 28.23 & 38.89 & 17.57\\
EAGLE   & DOTA  & 53.25 & 57.34 & 49.16\\
EAGLE   & EAGLE & 39.29 & 53.45 & 25.14\\%\cdashline{1-1}
\end{tabular}
\label{tab:Resultscrossvalidation}
\vspace{-.3cm}
\end{table}
}

\newcommand{\datasetCompStats}{%
\begin{table*}[t]
    \begin{center}
    \vspace{0.2cm}
    \caption{Comparison between EAGLE and datasets for object detection in aerial images. BB is short for bounding box. One-dot refers to annotations with only the center coordinates of an instance provided. Fine-grained categories are not taken into account. For example, EAGLE features 2 different categories with additional difficulty flags with respect to the class and orientation.}
    \resizebox{0.98\textwidth}{!}{
         \begin{tabular}{c|c c|| c c c c c c}
         \hline
         Datasets & \#~Vehicle  & \#~ Vehicle  & \#~All  & \#~Images  & \#~All  & Image  & Annotation & Year \\ 
                  & Instances   & Categories    & Categories&&Instances& Width (px)& Approach&
         \\
         \hline
         TAS~\cite{Heitz2008TAS}& 1,319 & 1 &  1 & 30 & 1,310 & 792 & HBB & 2008 \\
         NWPU-VHR‐10~\cite{Cheng2016NWOPU} & 232 & 1 & 10 & 800 & 3,775 & ~1000 & HBB & 2014\\
         VEDAI~\cite{Razakarivony2016VEDAI} & 3,270 & 6 & 9 & 1,210 & 3,640 & 1024 & OBB & 2015 \\
         UCAS‐AOD~\cite{Zhu2015UCAS} & 2,819 & 1 & 2 & 910 & 6,029 & 1280 & HBB & 2015 \\
         DLR-3K-Vehicle~\cite{Liu2015DLR3K} & 14,232 & 2 & 2 & 20 & 14,235 & 5616 & OBB & 2015 \\
         COWC~\cite{Mundhenk2016COWC} & 32,716 & 1 & 1 & 53 & 32,716 & 2000-19,000 & One-Dot & 2016 \\
         HRSC2016~\cite{Liu2016HRSC} & 0 & 0 & 1 & 1,070 & 2,976  & ~1000 & OBB & 2016 \\
         RSOD~\cite{Long2017RSOD} &0 &0 & 4 & 976 & 6,950  & ~1000 & HBB & 2017 \\
         DOTA~\cite{Xia2017DOTA} & 43,462 & 2& 15 & 2,806 & 188,282 & 300-4000 & RBB & 2017\\
         \hline
         \textbf{EAGLE~(ours)} & \textbf{215,986} & 2 & 2 & \textbf{8,280} & \textbf{215,986} & 936 & OBB & 2020\\
         \hline
         \end{tabular}
         }
    \label{tab:stats}
   \end{center}
    \vspace{-0.7cm}
\end{table*}
}

\newcommand{\DN}{EAGLE}

\begin{document}

\title{\DN: Large-scale Vehicle Detection Dataset in Real-World Scenarios using Aerial Imagery}
%\title{\DN: Large-scale Dataset for Oriented Vehicle Detection in Aerial Imagery in the Real-World Scenarios}

\author{\IEEEauthorblockN{Seyed~Majid~Azimi\IEEEauthorrefmark{1}\IEEEauthorrefmark{2}
,
Reza~Bahmanyar\IEEEauthorrefmark{1},
Corentin~Henry\IEEEauthorrefmark{1}, and
Franz~Kurz\IEEEauthorrefmark{1}}
\IEEEauthorblockA{\IEEEauthorrefmark{1}Remote Sensing Technology Institute, German Aerospace Center (DLR),
Wessling, Germany}%\\
%Emails: \{seyedmajid.azimi;~reza.bahmanyar;~corentin.henry;~franz.kurz\}@dlr.de}
\IEEEauthorblockA{\IEEEauthorrefmark{2}Department of Aerospace, Aeronautics and Geodesy, Technical University of Munich, Munich, Germany\\
%Emails: \{seyedmajid.azimi\}@dlr.de}
%Emails: \{seyedmajid.azimi;~reza.bahmanyar;~corentin.henry;~franz.kurz\}@dlr.de}
Corresponding author: \{seyedmajid.azimi\}@dlr.de}
}

\makeatletter
\let\@oldmaketitle\@maketitle
\renewcommand{\@maketitle}{\@oldmaketitle
\includegraphics[width=\linewidth,height=8\baselineskip]{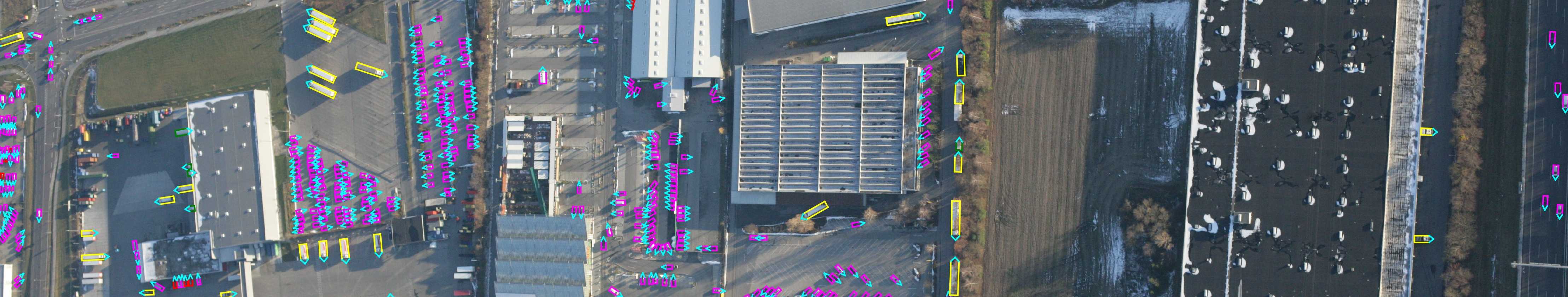}\\
\centering
    Sample aerial image from the EAGLE dataset with partial snow cover and its overlaid annotation taken in the early morning.\bigskip}
    \vspace{-3cm}
\makeatother
\vspace{-2cm}
\maketitle

\begin{abstract}
Multi-class vehicle detection from airborne imagery with orientation estimation is an important task in the near and remote vision domains with applications in traffic monitoring and disaster management. In the last decade, we have witnessed significant progress in object detection in ground imagery, but it is still in its infancy in airborne imagery, mostly due to the scarcity of diverse and large-scale datasets. Despite being a useful tool for different applications, current airborne datasets only partially reflect the challenges of real-world scenarios. To address this issue, we introduce EAGLE (oriEnted vehicle detection using Aerial imaGery in real-worLd scEnarios), a large-scale dataset for multi-class vehicle detection with object orientation information in aerial imagery. It features high-resolution aerial images composed of different real-world situations with a wide variety of camera sensor, resolution, flight altitude, weather, illumination, haze, shadow, time, city, country, occlusion, and camera angle. The annotation was done by airborne imagery experts with small- and large-vehicle classes. EAGLE contains 215,986 instances annotated with oriented bounding boxes defined by four points and orientation, making it by far the largest dataset to date in this task. It also supports researches on the haze and shadow removal as well as super-resolution and in-painting applications. We define three tasks: detection by (1) horizontal bounding boxes, (2) rotated bounding boxes, and (3) oriented bounding boxes. We carried out several experiments to evaluate several state-of-the-art methods in object detection on our dataset to form a baseline.  Experiments show that the EAGLE dataset accurately reflects real-world situations and correspondingly challenging applications. The dataset will be made publicly available.
\end{abstract}
\IEEEpeerreviewmaketitle
%%%%%%%%% BODY TEXT
\section{Introduction}
Automatic vehicle detection based on aerial imagery is crucial for a variety of applications such as large-scale traffic monitoring, parking lot utilization, urban planning, disaster management, as well as search and rescue missions. Aerial images, with their wide field of view% and small occlusion effects
, provide valuable information over large open areas in a short time~\cite{Ajay2017}.
\begin{figure*}
	\centering
	\vspace{0.3cm}
	\includegraphics[width=\textwidth]{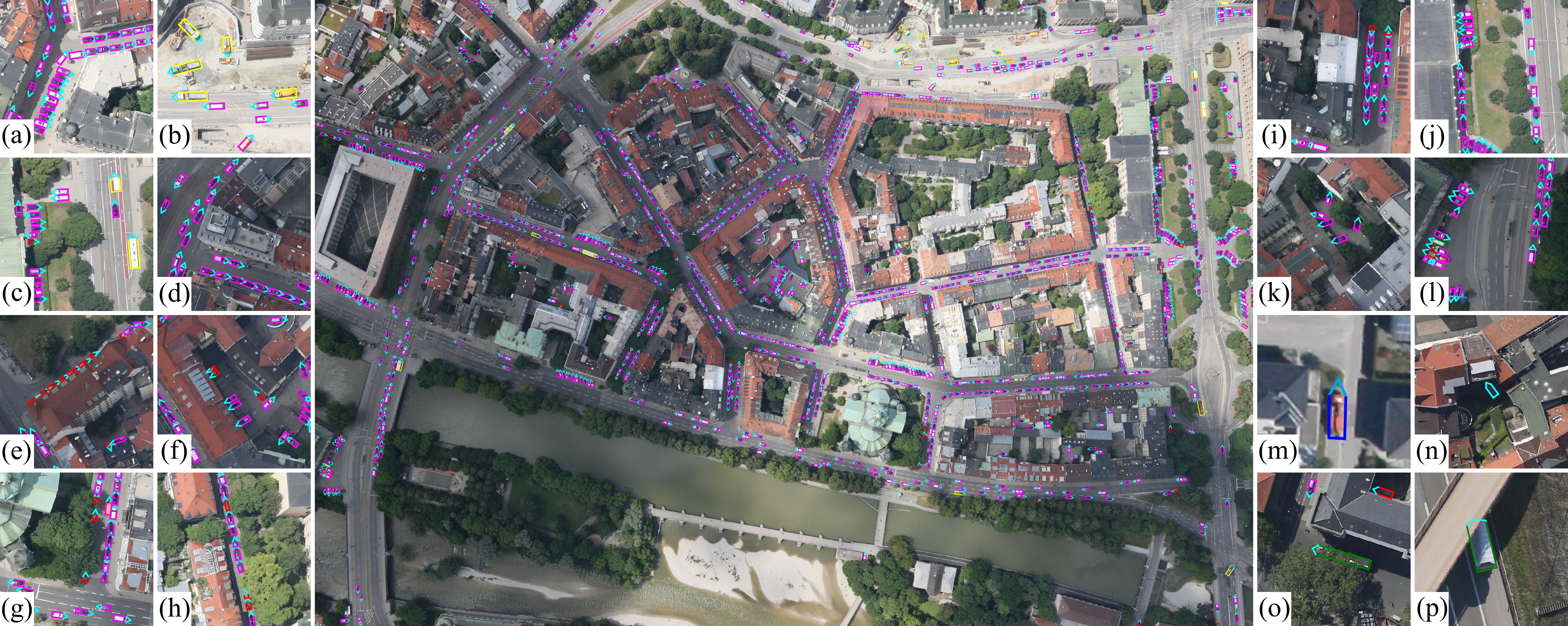}
    \caption{Sample annotations in EAGLE: (a-b) car and trucks in purple and yellow respectively, (c-d) sunny and cloudy illumination, (e-f) cars partly occluded by buildings, (g-h) cars partly occluded by vegetation, (i-j) cars in shadowed areas, (k-l) hard to identify cars orientations, (m) difficult car example, (n) car with weak orientation, (o-p) trucks with weak orientations.}
\label{fig:examples-big}
	\vspace{-0.6cm}
\end{figure*}
Due to the steep rise in the number of vehicles, traffic monitoring and management has become tremendously more complex, especially in urban areas. The major socio-economic impacts of the traffic-related problems such as air pollution, time loss in traffic jams, and health issues have increased the demand for developing novel automatic algorithms and adequate traffic data~\cite{Lewandowski2018}. It has been shown that vehicle detection algorithms based on aerial imagery can provide frequent and cost-efficient information about the location, number, and the types of vehicles in different traffic scenarios such as congestion caused by infrastructure bottleneck, accidents, or even lack of parking spaces~\cite{Ajay2017}. Due to the dynamic nature of traffic, the availability of large-scale information through aerial images can make traffic management more adaptive to the changing traffic conditions and help predicting infrastructure bottlenecks% for an effective infrastructure planning
~\cite{Souza2017}.
In disaster management, vehicle detection based on aerial imagery allows rapid localization of traffic congestion and abandoned vehicles to determine routes for effective search and rescue activities. Furthermore, in the case of natural disasters such as floods and earthquakes, aerial imagery is the most efficient means for detecting the affected vehicles% and save lives in vast affected areas
~\cite{Makiuchi2019}.
Recently, a large number of studies have focused on object detection (including vehicles) in aerial imagery~\cite{DBLP:journals/corr/PinheiroLCD16,recombinator,stacked,D-FPN}; however, despite the pronounced differences between ground and aerial images, most of the proposed methods are based on transferring object detection algorithms developed for natural-scene images to the aerial ones due to the scarcity of the large-scale aerial image datasets. 
For instance, to apply deep learning detection algorithms to aerial images, previous works usually relied on fine-tuning networks pre-trained on large-scale natural-scene datasets (\eg ImageNet~\cite{Deng2009}, MSCOCO~\cite{Lin2014}, PASCAL VOC~\cite{Everingham10thepascal}).
%
%Although these approaches show reasonable detection performance in remote sensing tasks, significantly improving their performance requires considering the specific properties of aerial imagery. 
%
As it can be seen in Figure~\ref{fig:examples-big}, the scale of the objects varies widely in aerial images due to not only the differences in spatial resolution
% of the sensors
, but also in the size of objects from the same category.
In addition, aerial images usually contain a large number of small objects distributed and oriented differently over the scene (\eg from sparse density of moving vehicles in highways to tightly packed ones in parking lots).
In addition, the number of the object instances in aerial images is unbalanced, from a few to thousands of objects per image.

Object detection in ground imagery owes its significant promotion to the large datasets such as MSCOCO, ImageNet, and PASCAL VOC. However, for aerial imagery, similar datasets in terms of image number and annotation details are scarce, which has highly limited the progress in developing methods for aerial images.% as well as multi-modality algorithms based on the fusion of natural scenes and aerial images.
The current available aerial image datasets \eg~\cite{Heitz2008TAS,Razakarivony2016VEDAI,Liu2015DLR3K,Zhu2015UCAS,Xia2017DOTA} suffer from either low number of images and annotated instances or low-quality annotations. 
%
%This is even worse in vehicle detection scenarios.
% DOTA
The largest currently available aerial image dataset for object detection is DOTA~\cite{Xia2017DOTA} which comprises 2,800 images with fifteen categories and about 188,000 bounding box annotations using already processed Google Earth and satellite images; however, it contains only 43,462 vehicles.
Other datasets such as TAS~\cite{Heitz2008TAS}, VEDAI~\cite{Razakarivony2016VEDAI}, COWO~\cite{Mundhenk2016COWC}, DLR-3K-Munich-Vehicle~\cite{Liu2015DLR3K}, and UCAS-AOD~\cite{Zhu2015UCAS} which mainly focus on vehicle detection also contain very limited number of annotated vehicles: TAS~(1,319), VEDAI~(3,270), COWO~(32,716), DLR-3K-Munich-Vehicle~(14,235), and UCAS-AOD~(2,819).
In addition to the number of instances, the inadequate diversity and complexity of the images used (\eg clear background and limited object distribution heterogeneity) in these datasets prevents them from representing real-world situations.
Table~\ref{tab:stats} shows detailed statistics from the current major aerial image datasets for object detection.
\datasetCompStats
To promote research on vehicle detection including vehicle detection, counting, and tracking% with their diverse applications, as mentioned before
, we propose a new and yet largest aerial image dataset for vehicle detection in real-world aerial imagery scenarios, called \gls{EAGLE}. %This dataset is composed of \todo{345 high-resolution aerial images of size \SI{5616x3744}{\px}} acquired by commercially available standard DSLR cameras installed on an airborne platform. The images were taken 
%from different flight campaigns 
%over various traffic scenarios (\eg motorways, urban/rural areas, and industrial district), in different seasons, and different weather conditions in five countries. %Germany and four other European countries.
%It includes real-world disaster situations such as flood, landslide, and earthquake.
%A group of experts% in aerial imagery interpretation 
%annotated the images for two common vehicle categories, namely small and large vehicles.
%, with oriented bounding boxes
%% (to allow for the detection of arbitrary vehicle orientations)
%, resulting in 215,986 instances, which is 5$\times$ more than the number of vehicle instances in
% DOTA, 
%the largest existing dataset% for vehicle detection 
% to date.

Altogether, the main contributions of this paper are:
\begin{itemize}
    \item EAGLE, which is to the best of our knowledge the largest aerial image dataset for vehicle detection and the first dataset of its kind addressing real-world scenarios. 
    \item Its high-quality annotations can contribute to the development and evaluation of practical airborne vehicle detection systems as well as haze, shadow, in-painting and super-resolution applciations. %We keep updating EAGLE by increasing the number and diversity of images. 
    %The dataset will be made publicly available.
    \item Benchmarks of state-of-the-art object detection algorithms as baseline for future works by defining benchmarks for all three possible detection possibilities and two dataset split approaches.
\end{itemize}
\begin{figure*}[t!]
    \vspace{0.2cm}
    \includegraphics[width=.25\textwidth]{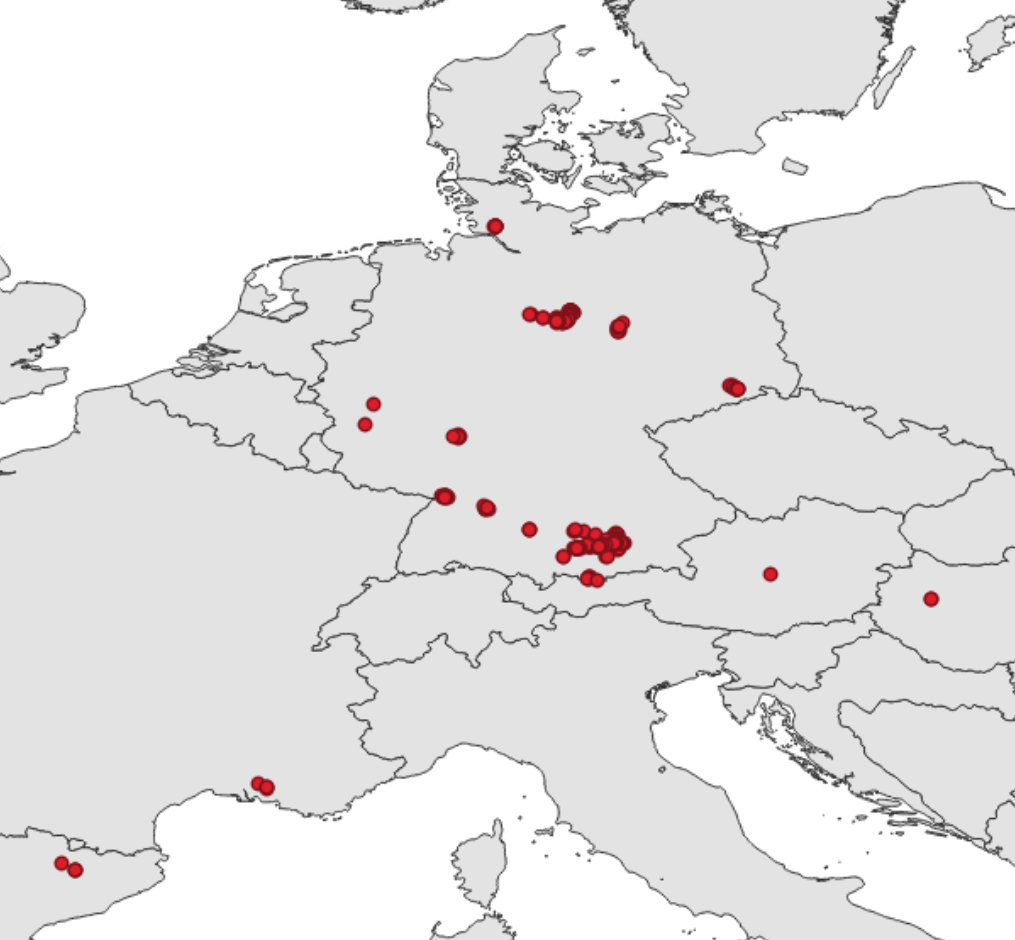}
    \includegraphics[width=0.75\textwidth]{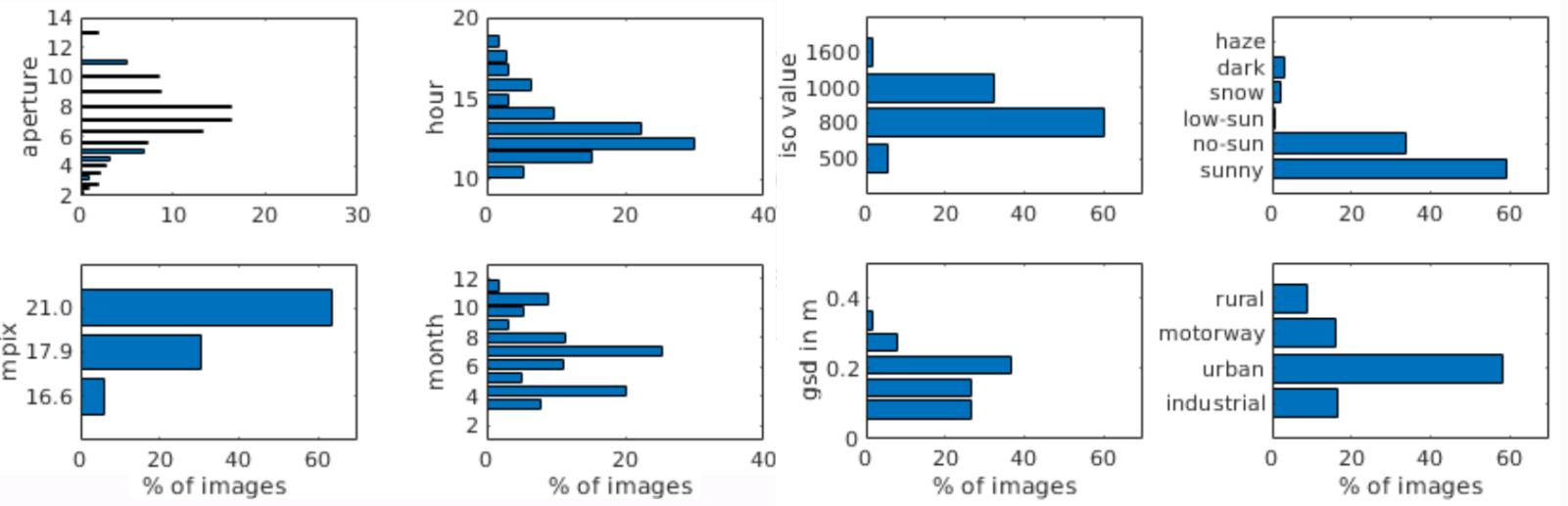} 
    \caption{Distribution of image acquisition locations over central Europe, as well as the statistics on camera parameters, image, and scenery properties.}
	\label{fig:dataset2}
	\vspace{-0.45cm}
\end{figure*}
%%%%%%%%% BODY TEXT
\section{EAGLE dataset}
\begin{figure}[t!]
 	\centering
 	\includegraphics[width=0.45\textwidth]{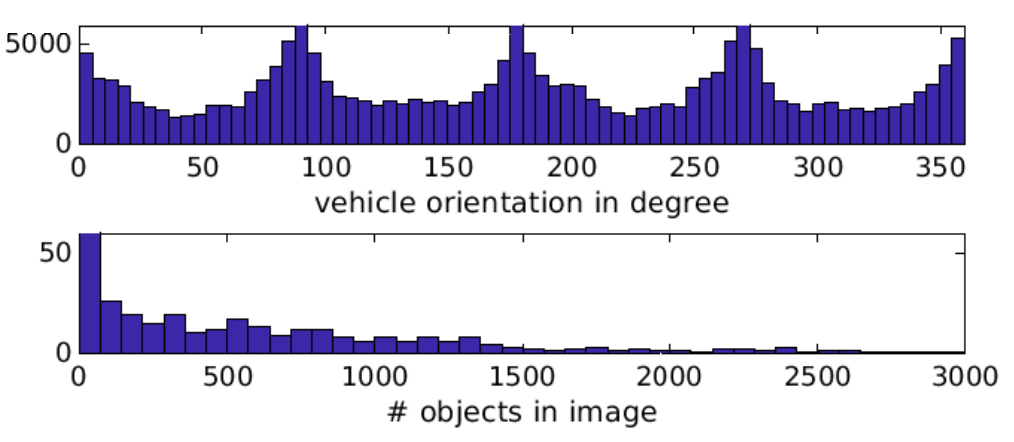} 
 	\caption{Statistics of annotated vehicles with respect to vehicle orientation (top) and instances per image (bottom).}
 	\label{fig:examples-hist}
 	\vspace{-0.5cm}
 \end{figure}

\begin{figure*}[t!]
    \vspace{0.3cm}
	\centering
	\begin{tabular}{cccc}
	\includegraphics[width=0.22\textwidth]{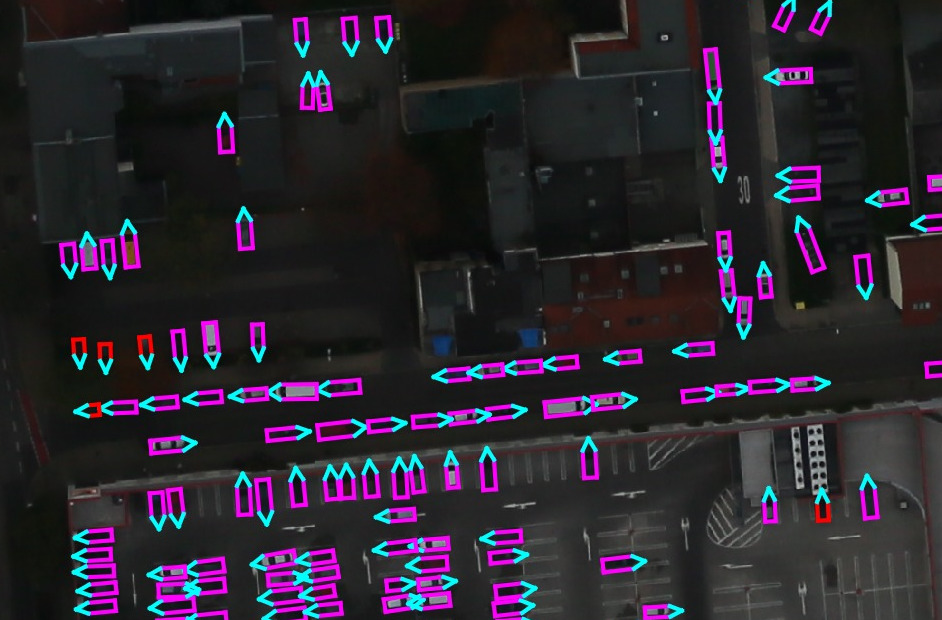} & 	\includegraphics[width=0.22\textwidth,trim={0 0 0 120},clip]{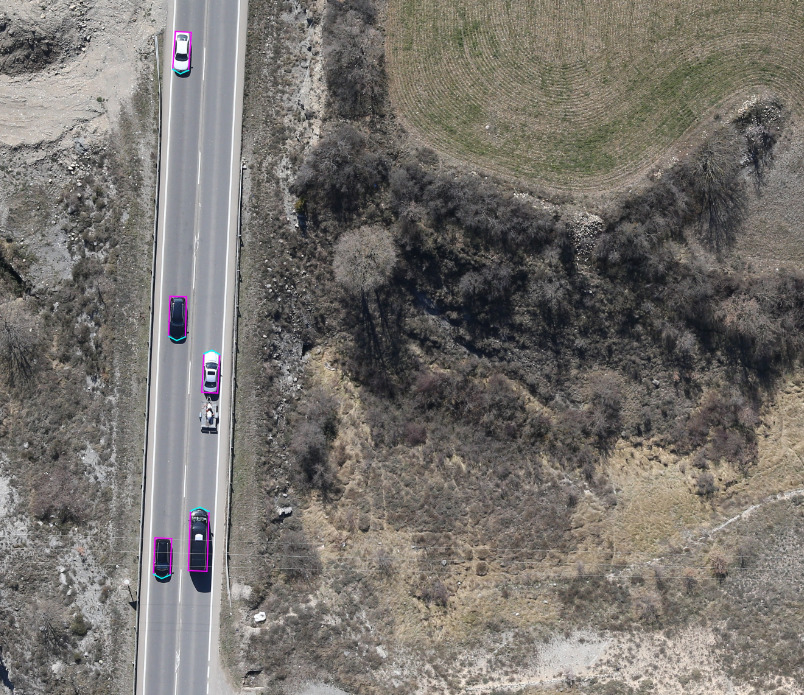} & 	\includegraphics[width=0.22\textwidth,trim={0 0 0 30},clip]{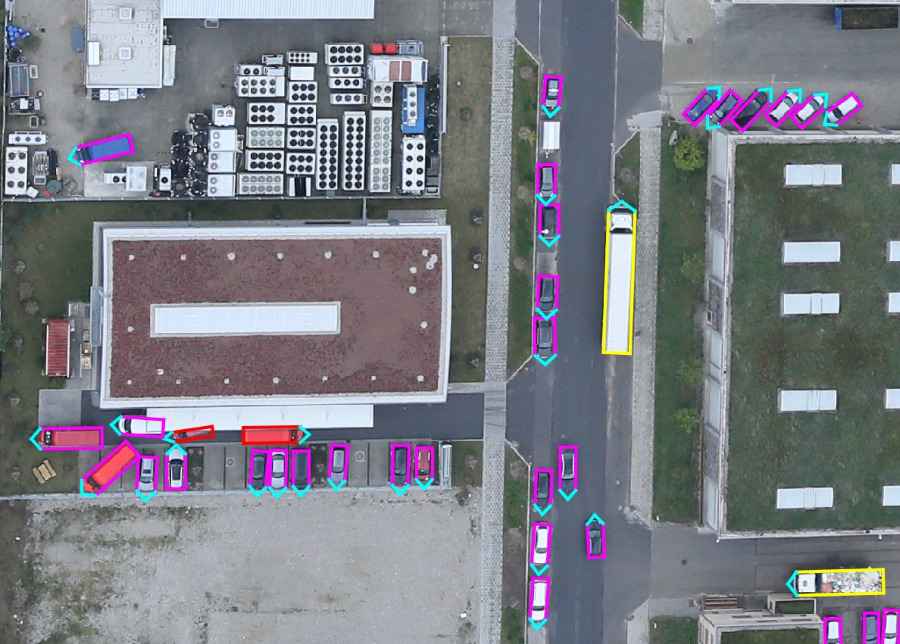} & 
	\includegraphics[width=0.22\textwidth,trim={0 0 0 40},clip]{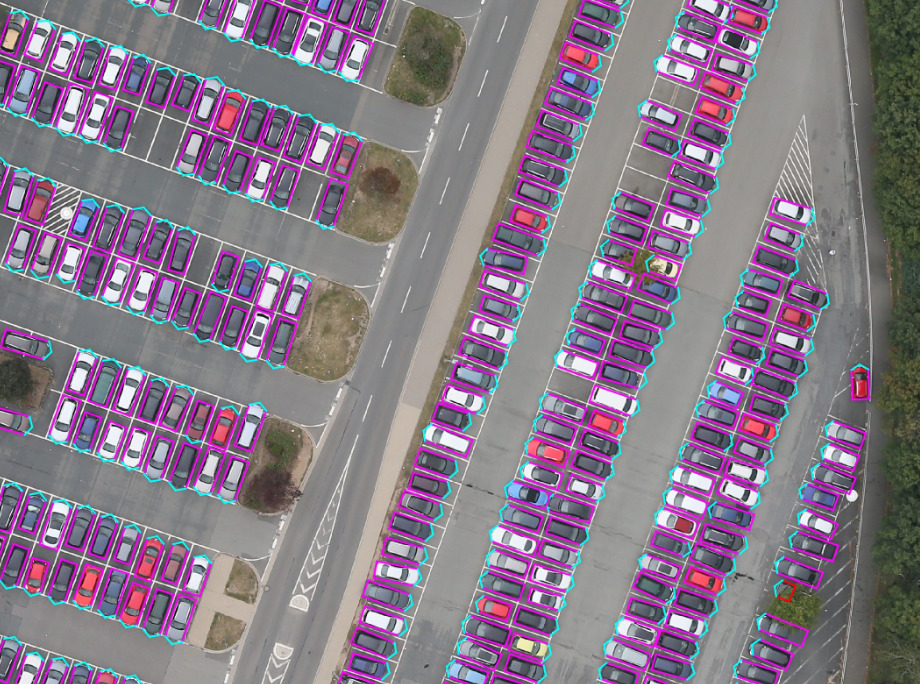} \\	
	\includegraphics[width=0.22\textwidth]{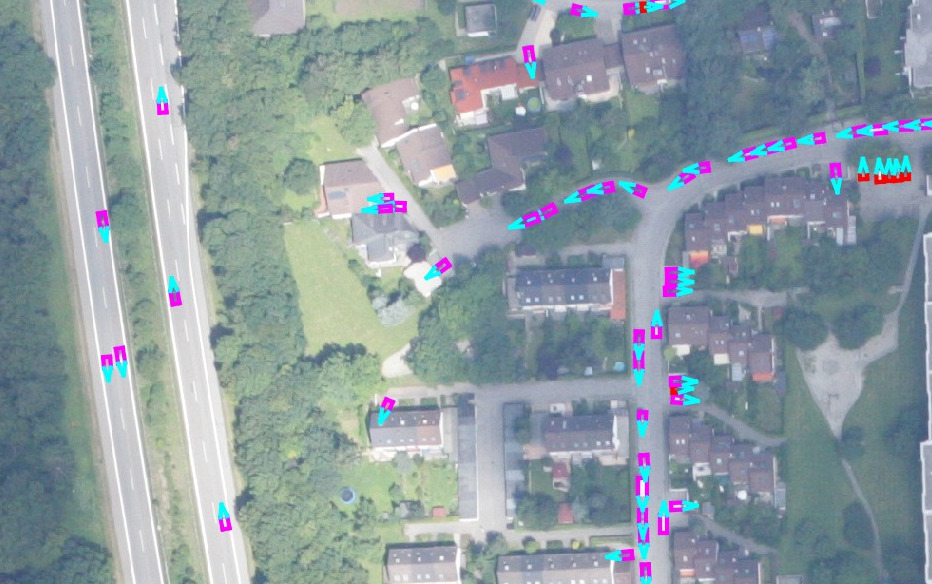} & 	\includegraphics[width=0.22\textwidth]{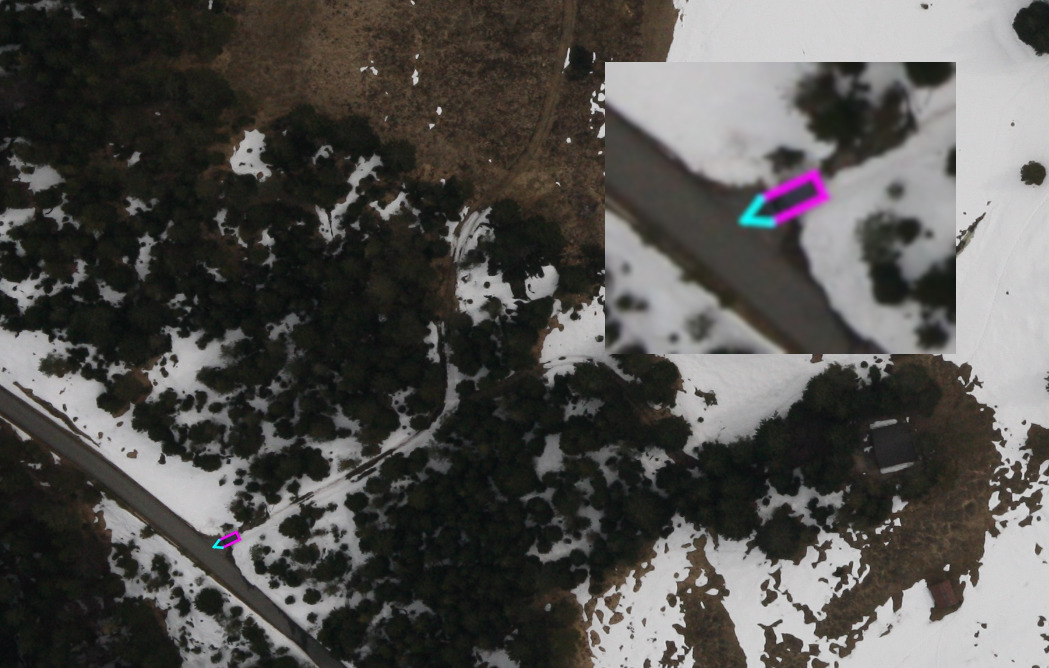} & 	\includegraphics[width=0.22\textwidth,trim={0 0 0 75},clip]{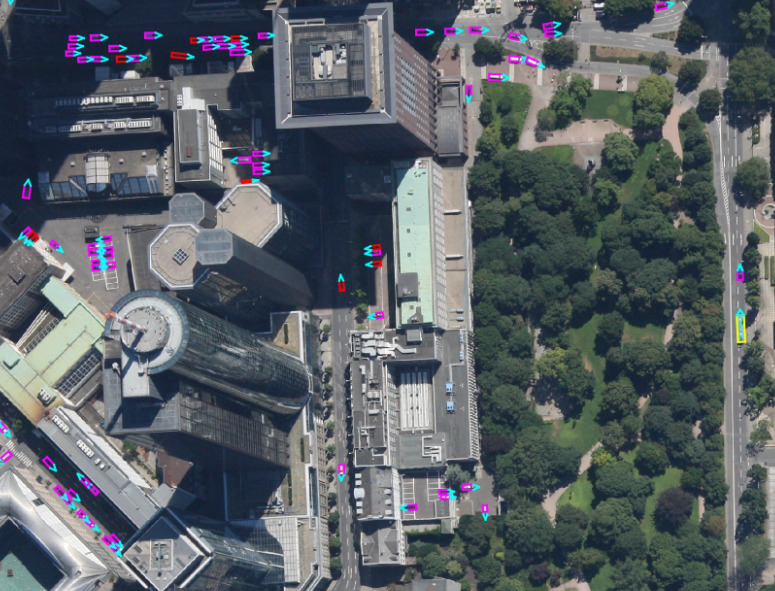} & 
	\includegraphics[width=0.22\textwidth]{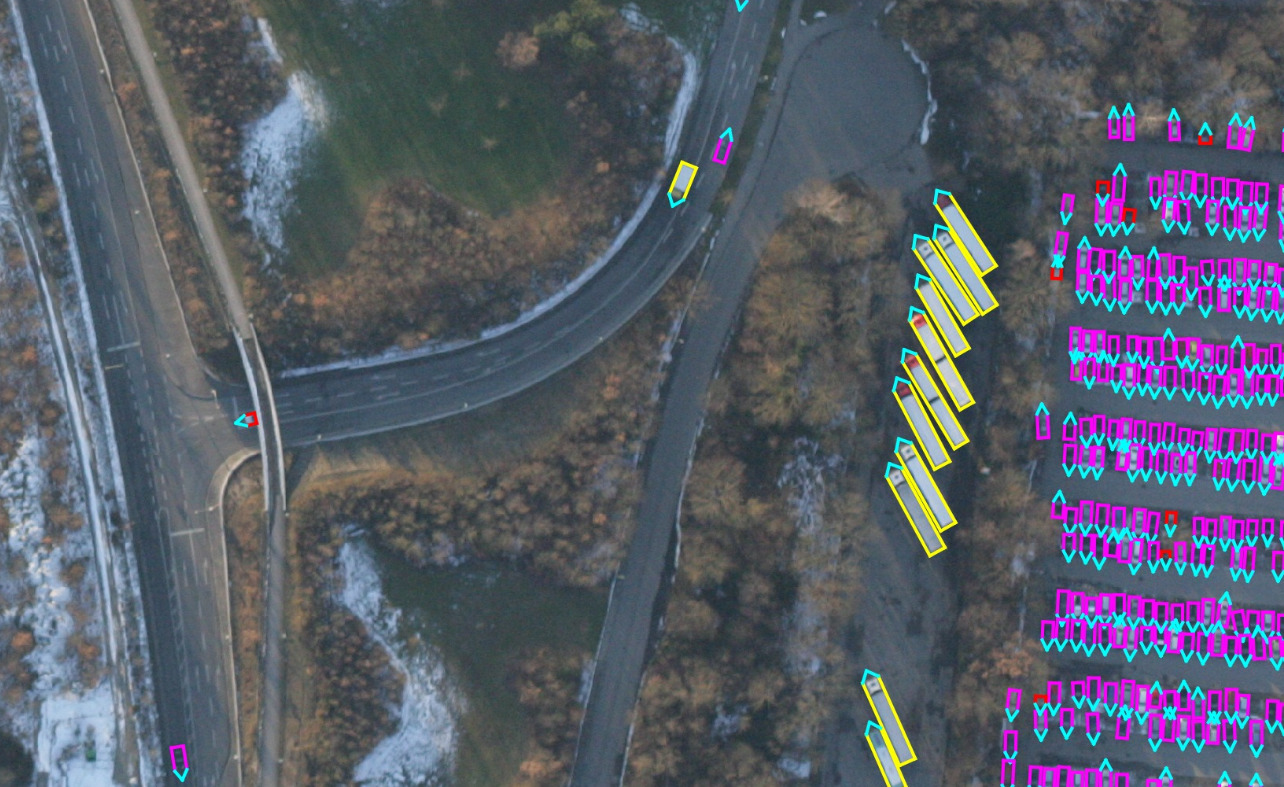} \\	
	\includegraphics[width=0.22\textwidth]{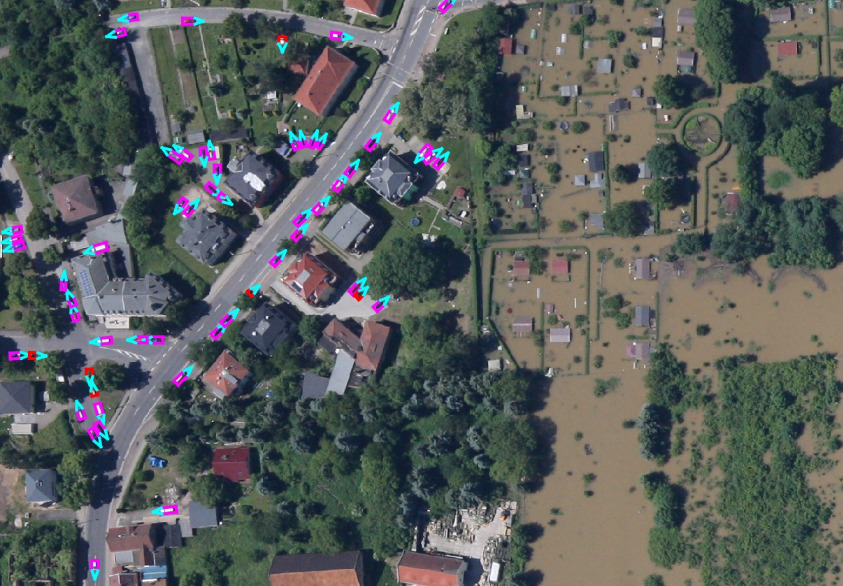} & 	
	\includegraphics[width=0.22\textwidth]{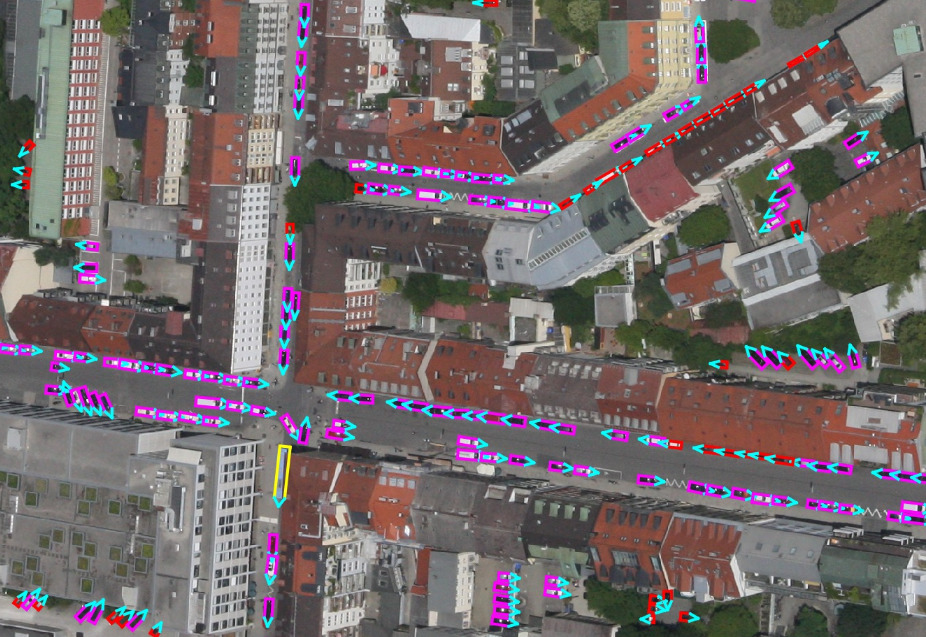} & 	
	\includegraphics[width=0.22\textwidth,trim={0 0 0 65},clip]{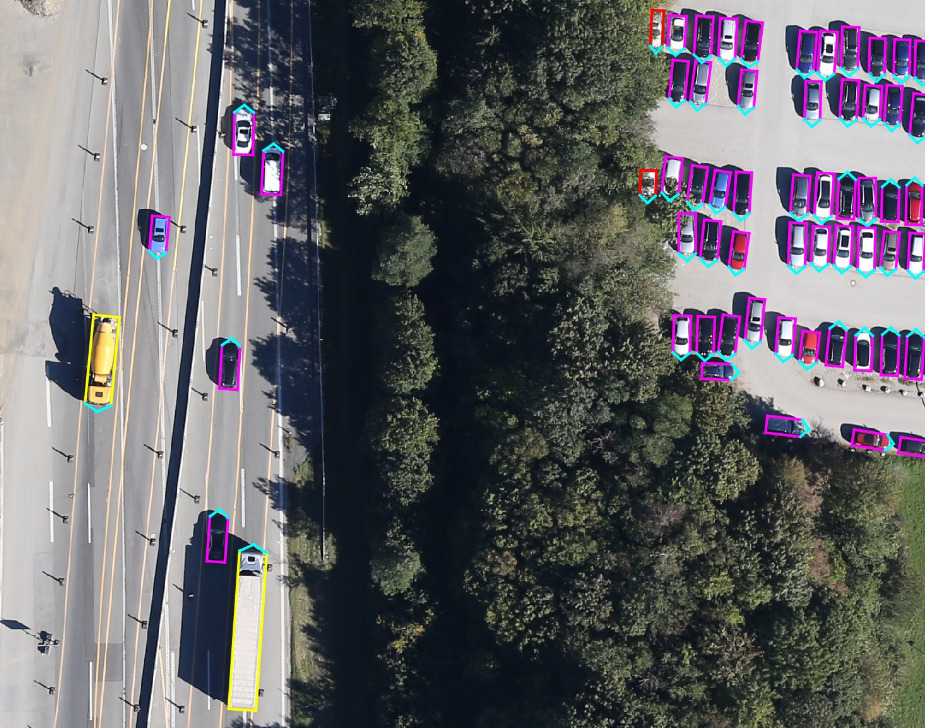} & 	
	\includegraphics[width=0.22\textwidth,trim={0 0 0 35},clip]{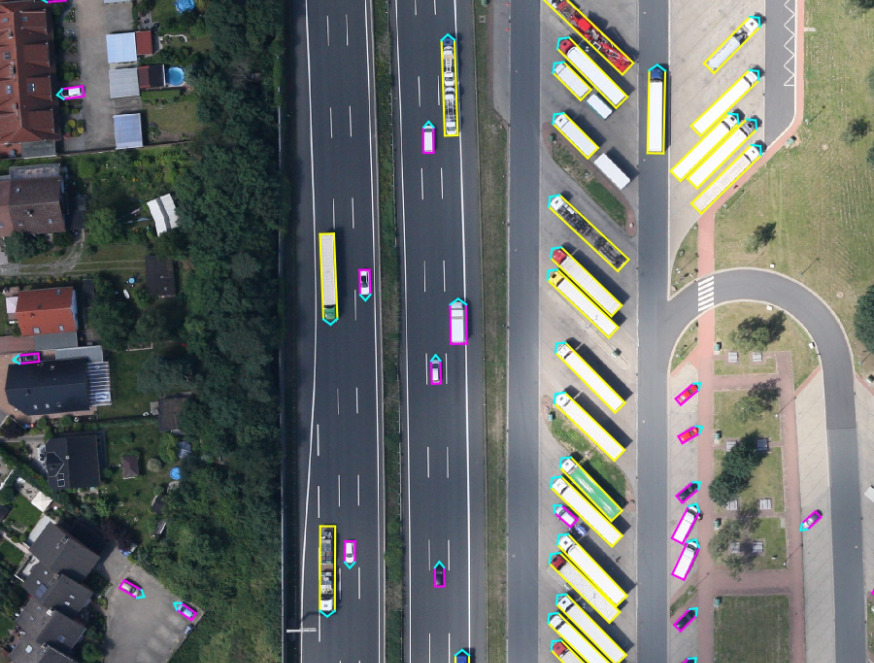} \\	
	\includegraphics[width=0.22\textwidth]{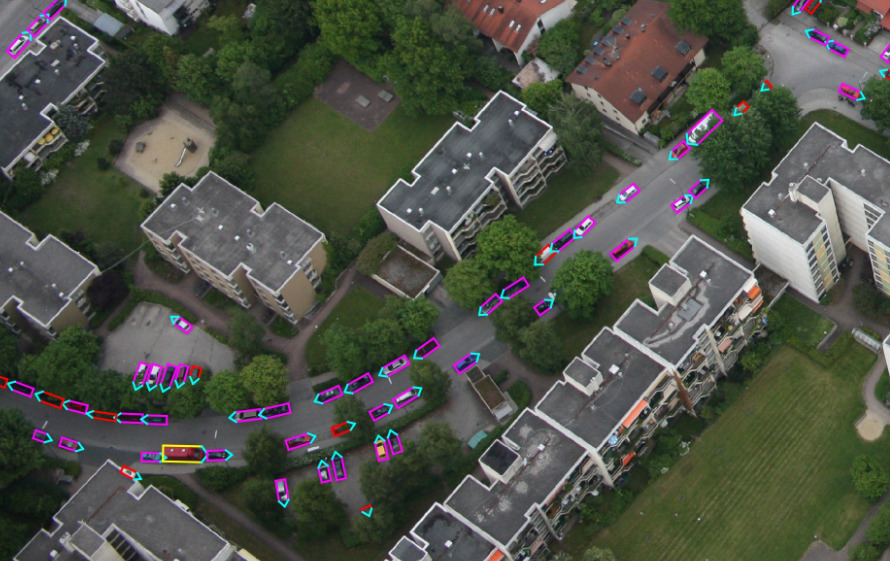} & 	
	\includegraphics[width=0.22\textwidth,trim={0 0 0 65},clip]{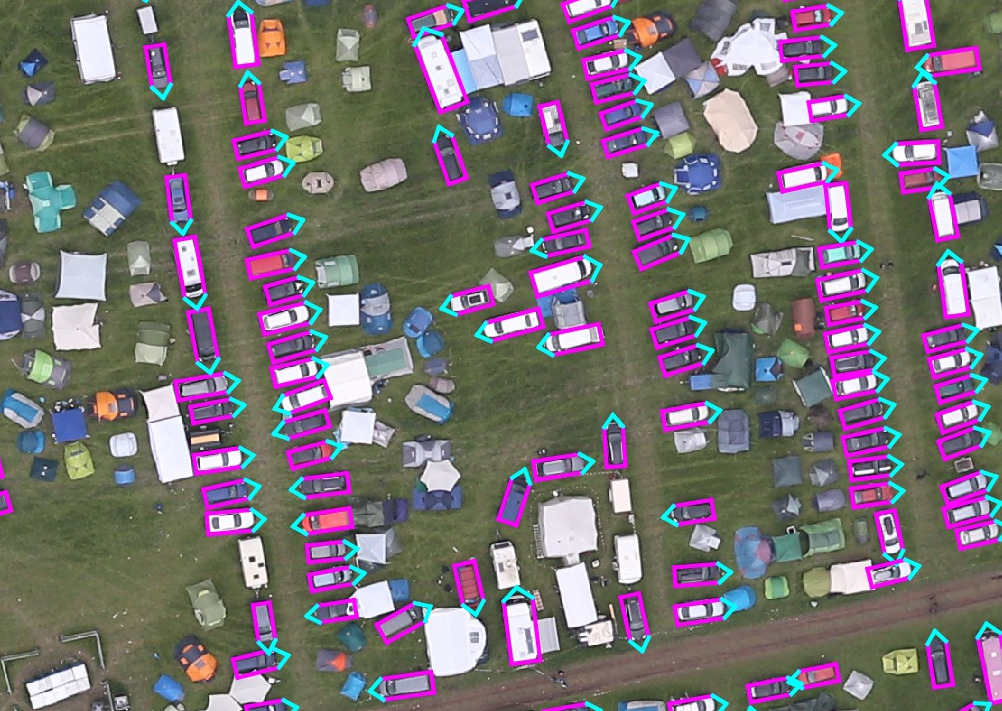} & 	\includegraphics[width=0.22\textwidth,trim={0 0 0 15},clip]{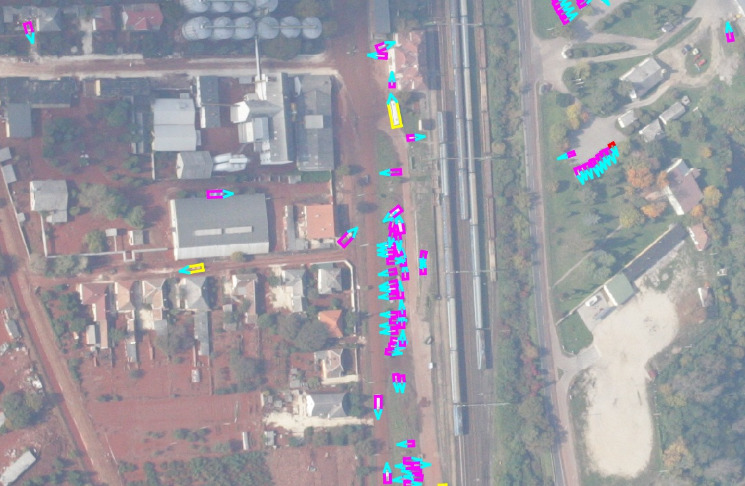}& 
	\includegraphics[width=0.22\textwidth,trim={0 0 0 10},clip]{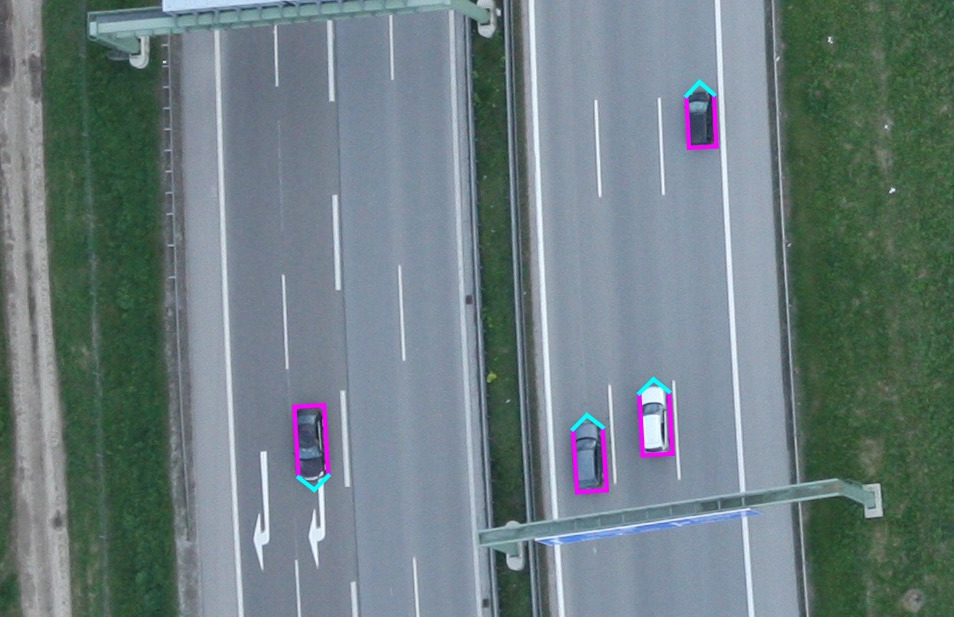} \\	
    \end{tabular}
	\vspace{-0.1cm}
	\caption{Examples of annotated images (left to right, top to bottom): low sun, rural scene, industrial scene, parking space, mixed illumination, snow, mega city, mixed parking, flood scene, oblique view, highway scene, service area, suburban area, festival scene, haze, motorway. Magenta: small vehicles. Yellow: large vehicles. Cyan triangle: driving direction.}
	\label{fig:examples}
	\vspace{-0.5cm}
\end{figure*}

The~\gls{EAGLE} dataset consists of $8,820$ aerial images with size of~\SI{936x936}{\px},
% I have decided to show number of sub part of original images as it seems no one pays attention to the images size, but rather image number. Below, I have shortly mentioned that in case images are stitched, the number of images will be 345. Each 3K image contains 24 936x936 image.
% \todo{$345$ large RGB aerial images with sizes of \SI{5616x3744}{\px}}
acquired during several flight campaigns carried out 
between 2006 and 2019 in various time of day and year with different weather and illumination conditions. The images were taken under different traffic conditions and situations involving vehicles such as motorways, urban/rural areas, industrial districts,
floods, wildfires, earthquakes, as well as search and rescue missions over multiple locations in five 
%European
countries % including Germany, Austria, Hungary, France, and Spain
(see Figure~\ref{fig:dataset2}). The images contain a large diversity of vehicle orientation angle and number of objects per image as shown in Figure~\ref{fig:examples-hist} with a higher number of vehicle instances compared to previous datasets (see Figure~\ref{fig:categ_comp}). Figure~\ref{fig:examples} showcases some example image patches from the dataset. We acquired the images using a camera system comprised of three standard DSLR cameras (Canon EOS cameras) mounted on an airborne platform with different looking angles, a nadir-looking (top-down vertical) and two side-looking cameras.
According to the conditions of the flight campaigns, the camera setups such as aperture size, image size, and ISO were adjusted differently. 
The platform was installed either on an airplane or on a helicopter flying at altitudes between \SI{300}{\m} and \SI{3000}{\m}, resulting in a range of\glspl{gls:GSD}, or spatial resolution, from \SI{5}{\cm} to \SI{45}{\cm} per pixel.
The images were taken from early in the morning until the evening in various weather conditions (\eg sunny, snowy, rainy, and foggy) with different illumination levels.
%The precise image geographical information (location coordinates) and acquisition time were also available, which helped us to select diverse and non-duplicated images.
Altogether, the variability in image parameters and scenes allows our dataset to cover a wide range of real-world situations involving vehicles.
Figure~\ref{fig:dataset2} represents further statistics on the %image properties of the 
EAGLE dataset.
%
%Most of the images were acquired with an aperture size of $f/8$, an image size of $21$ megapixels, an ISO of $800$, and a shutter speed of $1/2000$~s. In addition, the mean\gls{gls:GSD} of the images is around \SI{15}{\cm}/px.
\subsection{Image annotation}\label{sec:imageannotation}
\begin{figure}[t!]
    \centering
    \vspace{0.2cm}
    \includegraphics[width=0.45\textwidth]{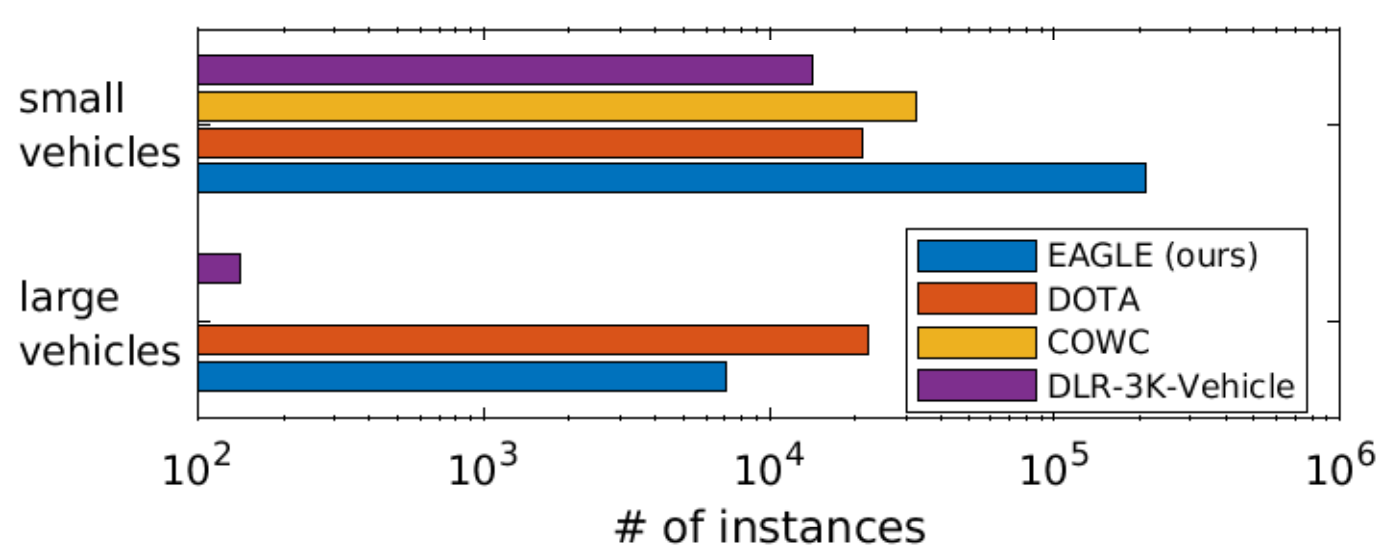}
    \vspace{-0.1cm}
    \caption{Comparison between the number of annotated small and large vehicles in the EAGLE dataset and the vehicle sets of other aerial image datasets.}
    \label{fig:categ_comp}
    \vspace{-0.5cm}
\end{figure}

Taking into account the relevance of the vehicle categories for the real-world applications of aerial imagery according to experts in the domain, we decided on two main categories for our dataset, namely small vehicles (cars, vans, transporters, SUVs, ambulances, police cars) and large vehicles (trucks, large-trucks, minibuses, buses, firefighter trucks, construction vehicles, trailers).
The annotation contains the coordinates of all four vehicles corners having right angle between sides as well as orientation degree between \ang{0} to \ang{360} indicating the angle of vehicle head with respect to the trigonometric circle.
Table~\ref{tab:stats} shows 
%statistics on the EAGLE dataset and 
a comparison between EAGLE and other existing aerial imagery datasets for vehicle detection. The EAGLE 
%dataset was manually annotated for two main vehicle classes (small and large vehicles) with oriented bounding boxes, which resulted in 
contains $215,986$ annotated vehicles, ranging from 1 to 3,567 annotations per image in all possible orientations (see Table \ref{tab:general}), making it the largest aerial image dataset for vehicle detection by a large margin (5$\times$ more vehicle instances than in the current largest dataset). Furthermore, for each instance, the visibility condition (totally/partly/hardly visible) and orientation clarity (clear/unclear) of the vehicle were provided.
%%%%%%%%%%%%%%%%%%%%%%%%%%%%%%%%%%%%%%%%%%%%%%%%%%%%%%%%%%%%%%%%%%%%%%%%%%%%%
Stitched images with original sizes are $345$ ones of \SI{5616x3744}{\px} size.
%%%%%%%%%%%%%%%%%%%%%%%%%%%%%%%%%%%%%%%%%%%%%%%%%%%%%%%%%%%%%%%%%%%%%%%%%%%%%
%
%While a number of existing datasets~\cite{Heitz2008TAS,Cheng2016NWOPU,Zhu2015UCAS,Mundhenk2016COWC} consider a single category for all different vehicle types, we believe that the small and large vehicles should be assigned to different categories due to their considerably different appearances and influences on the traffic scenarios. 
As visible in Table~\ref{tab:general}, the EAGLE dataset contains 208,963 small and 7,023 large vehicles. A category-wise comparison  %between our dataset and the vehicle sets of the existing aerial image datasets 
is provided in Figure~\ref{fig:categ_comp}. 
%Although the number of large vehicles in EAGLE is slightly smaller than that of the DOTA dataset, the number of small vehicles significantly surpasses all existing datasets ($\sim10\times$ more than DOTA).

\subsection{Annotation method}
We have addressed various challenges during the annotation of the vehicles in our aerial images. Due to the diversity of the scene locations, the acquisition time, as well as the weather and illumination conditions, precise annotation of the vehicles could be a very challenging task. For example, in an image taken over a flooded area when haze is present with low illumination or resolution, the visibility of the vehicles gets limited considerably. In addition, the occlusion due to other objects or strong shadow could cause difficulties in finding the vehicles. Furthermore, spotting vehicles in large aerial images of remote places (\eg mountains) is not trivial. Moreover, categorizing the vehicles into either small or large vehicles could be sometimes tricky due to the uncertainty about the category of some borderline cases such as large transporters or buses. To ease the latter situation, we assumed the one-cabin vehicles with a width or a height smaller than a specific threshold (specified by an expert) as small vehicles and otherwise as large vehicles. We also assigned a difficulty flag for the occluded vehicles which can help to better train algorithms to overcome occlusion. Detecting the occluded vehicles is very important in real-world scenarios such as in disasters like flood when the vehicles are trapped or partially under water.%, as lives are at stake.
% For the annotation of the EAGLE dataset, we relied on a bounding box approach, the so-called~\gls{gls:OBB}. In order to ensure the annotation quality, a special care was taken for the precision of the bounding boxes so that they would tightly correspond to the vehicle contours. Bounding box imprecision could cause major drawbacks in the detection performance, particularly in the oblique view images.
%
In the ground imagery % or near vision
, objects are usually annotated by\gls{gls:HBB}, where an HBB can be defined by its top-left (TL) and bottom-right (BR) vertices, $(x_{TL}, y_{TL},x_{BR}, y_{BR})$; or by its center point $(x_c, y_c)$ together with the width $w$ and height $h$, ($x_c,y_c,w,h$).
HBB is an efficient object annotation approach; however, it does not consider the objects' orientation, which can lead to imprecise outlines of arbitrary oriented objects.% such as those in aerial images.
Moreover,\glspl{gls:HBB} considerably overlap when objects are tightly packed, which can confuse even state-of-the-art algorithms trying to distinguish them.
\begin{table}[t!]
	\centering
	\resizebox{.75\columnwidth}{!}{\begin{tabular}[b]{l||c|c}
	    & Small vehicles  & Large vehicles \\\hline
    	\# Annotations & 208,963 & 7,023 \\\hline
    	\# Weak orientation & 311 & 10 \\\hline
    	\# Partly visible &  18,188 & 184  \\\hline
        %min/max/avg & \multirow{1/3,567/630} & \multirow{0/140/16}\\ 
        min/max/avg & 1/3,567/630 & 0/140/16\\ 
        objects per image & & \\\hline
    \end{tabular}}
	\caption{Category-wise statistics in \gls{EAGLE}.}
	\label{tab:general}
	\vspace{-0.5cm}
\end{table}
An approach toward alleviating the limitations of HBB is using arbitrary quadrilateral bounding boxes, the so-called\gls{gls:RBB}~\cite{Xia2017DOTA}, which can be described by $\{(x_i, y_i),~i= 1,2,3,4\}$, where $(x_i, y_i)$ are the vertex coordinates which can be with a clockwise order~\cite{Xia2017DOTA}. A specific case is a rotated rectangle when the sides make right angle with each other.
%
%RBB can be extended to OBB by to consider the object orientations through arranging the vertices in a clockwise order.
%
%Although OBB improves the precision of the object outlines, it is not constrained for generating bounding box with right angles which can introduce noise to the annotations that require such bounding boxes (\eg vehicles.) 
%
Inspired by~\cite{shi2017detecting,Xia2017DOTA} and the annotations in the common object detection benchmarks such as MSCOCO and PASCAL VOC, we propose a right-angle constrained\gls{gls:OBB} which can be described as $\{(x_i, y_i),~i= 1,2,3,4;\theta \}$, where $(x_i, y_i)$ are the vertex coordinates %(similar to~\cite{shi2017detecting,Xia2017DOTA}) 
and $\theta$ indicates the bounding box orientation. \gls{gls:OBB} can be also represented as $(x_c, y_c, w, h, \theta)$, where the bounding box edges are oriented according to $\theta$. This approach ensures the precision of the object outlines. %, but also speeds up the annotation process to a large degree compared to the arbitrary quadrilaterals. 
%We employ a custom tool implementing our proposed OBB approach for annotating the EAGLE dataset. In this tool, the annotator only needs to place the rear middle point of each vehicle and move the cursor along the vehicle direction up to its front end, and then adjust the width. This tool also allows generating\gls{gls:HBB} and\gls{gls:RBB} annotations. 
%Some image patches are shown in Figure~\ref{fig:examples-big} as example annotations.
\subsection{Dataset splits}
We split the dataset into training, validation, and test sets based on two approaches. In the first approach, we randomly assign $1/2$, $1/6$, and $1/3$ of the images respectively.
In this case, images from similar %acquisition missions or 
flight campaigns can be present in both train and test sets, which makes the detection task easier and similar to DOTA.%, and therefore the performance of the algorithms would be higher. This is one reason why the state-of-the-art on DOTA has increased from $50\%$~mAP to $80\%$~mAP in a short period of time. 
Thus, in the second approach, we split the dataset so that the images from the same flight campaigns are either in the training or test set. This approach is similar to the real-world scenarios in which there is no prior knowledge about future flight missions and their locations, weather or illumination conditions.
%
%This approach introduces new challenges to the detection task and evaluates the generalization ability of the algorithms to unforeseen detection scenarios.
%
%We will make the\gls{EAGLE} dataset publicly available. %with all images in their original size and the ground truth for the training and validation sets, while the ground truth of the test set will remain undisclosed. The detection results will be evaluated on our online platform.
\subsection{Contributions over the existing datasets}
%EAGLE is the first large-scale aerial vehicle dataset comprised of images acquired between 2003 and 2019 within 5 different countries by different camera sensors and configurations with resolutions ranging between 5cm/px to 30cm/px. 

\begin{figure}[t!]
\centering
\vspace{0.2cm}
\subfigure{\includegraphics[width=3.9cm,height=1.6cm,angle=90]{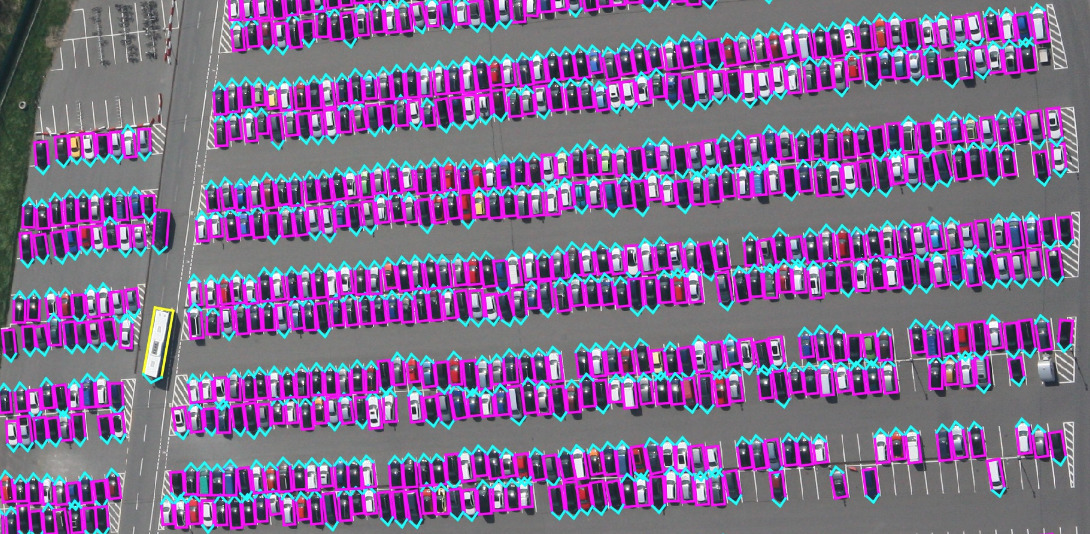}}
\subfigure{\includegraphics[width=3.9cm,height=1.4cm,angle=90]{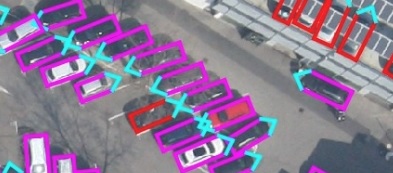}}
\subfigure{\includegraphics[width=3.9cm,height=.6cm,angle=90]{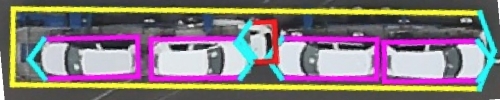}}
 \hspace{1.5em}
\subfigure{\includegraphics[width=3.9cm,height=1.6cm,angle=90]{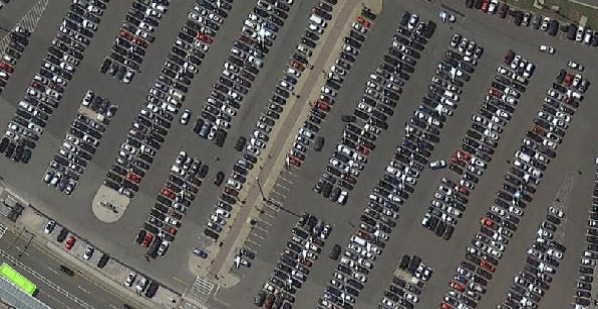}}
\subfigure{\includegraphics[width=3.9cm,height=1.4cm,angle=90]{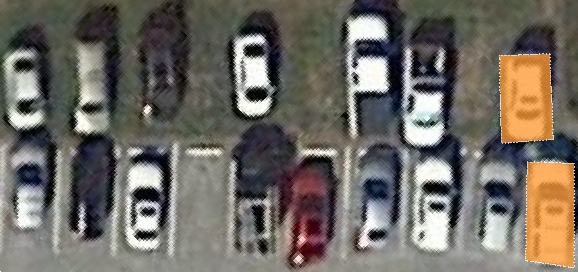}}
\subfigure{\includegraphics[width=3.9cm,height=.6cm,angle=90]{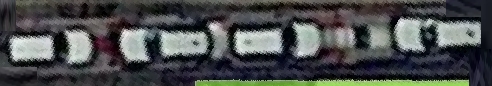}}
	\caption{High-quality EAGLE labels (left column) and incomplete DOTA labels (right column).}
	\label{fig:dota-eagle}
	\vspace{-.6cm}
\end{figure}

%EAGLE aims at solving the vehicle detection problem in aerial imagery for real-world situations in which the flight campaigns could drastically differ in location, illumination and weather conditions. 
The existing datasets containing vehicle instances (e.g. DOTA) suffer from inconsistent or inaccurate annotations, low degree of diversity and a small number of vehicle instances, %(DOTA has $5\times$ less than EAGLE), 
limiting their practical applications. Therefore, vehicle detection datasets such as EAGLE with thorough annotations even for tiny yet visible vehicles (see Figure~\ref{fig:dota-eagle}) are lacking in the community.
Moreover, EAGLE enables researchers to do research on haze and shadow removal as well as super-resolution, in-paining and instance segmentation.
Our dataset is featuring major differences compared to the DOTA dataset:
\begin{itemize}
    \item EAGLE focuses on vehicle detection in real-world and practical scenarios  %(e.g. traffic monitoring and disaster management) 
    with images of diverse location, time, resolution, weather and illumination conditions while DOTA is a multi-class general-purpose detection and classification dataset. %comprised of cloud- and haze-free aerial and satellite images
    %,   
    %A model trained on EAGLE will therefore more easily generalize to images taken under adverse conditions, for instance after a natural disaster.
    \item DOTA suffers from incomplete and noisy annotations (see Figure~\ref{fig:dota-eagle}) especially for small vehicles~\cite{azimi2018towards}, whereas EAGLE provides precise and comprehensive annotations (even for partially visible vehicles).
    %\item For traffic monitoring tasks, object localization is more important than categories exhaustiveness. Therefore, and contrary to DOTA, EAGLE purposely provides a limited set of accurately labeled vehicle classes.
    %\item The annotations of EAGLE include vehicle orientation, a valuable information for vehicle tracking tasks. We also introduced a CNN which uses the vehicle orientation together with the bounding box coordinates for training. Results showed more accurate predictions compared to using only bounding box coordinates. 
    \item Due to overlaps between the training and test sets in DOTA, the task is less challenging than EAGLE in which two training/test splits are proposed: (1) a random patch-based split, and (2) a more realistic and challenging campaign-based split, where the test set contains locations and adverse conditions unseen during training.
\end{itemize}
%%%%%%%%% BODY TEXT
\section{Evaluation}
We assess the performance of state-of-the-art object detection methods on\gls{EAGLE}.
For\gls{gls:HBB} object detection, we choose 
Cascade (Mask-)RCNN~\cite{cai2019cascade},%\footnote{\label{note3}https://github.com/open-mmlab/mmdetection},
% ~Cascade RCNN~\cite{cai2018cascade},%\footnote{https://github.com/zhaoweicai/cascade-rcnn}, 
~Mask-RCNN~\cite{he2017maskrcnn}\footnote{https://github.com/facebookresearch/Detectron\label{fn:note1}}, 
FPN~\cite{fpn}, %\footnotemark[\ref{fn:note1}], 
Faster RCNN~\cite{fasterrcnnNIPS2015}, %\footnotemark[\ref{fn:note1}], %GHM~\cite{li2019gradient}\footnote{https://github.com/libuyu/GHM\_Detection}, 
%Mask Scoring R-CNN~\cite{huang2019mask}\footnote{https://github.com/zjhuang22/maskscoring_rcnn}, 
FCOS~\cite{tian2019fcos}\footnote{https://github.com/tianzhi0549/FCOS}, 
%Double-Head R-CNN~\cite{huang2019mask}\footnotemark[\ref{note3}], 
%GCNET~\cite{cao2019gcnet}\footnote{https://github.com/xvjiarui/GCNet},fn:
%Grid R-CNN (Plus)~\cite{lu2019grid}\footnote{https://github.com/STVIR/Grid-R-CNN}, 
%Hybrid Task Cascade~\cite{chen2019hybrid}\footnotemark[\ref{note3}], 
%HRNet~\cite{sun2019deep,sun2019high}\footnotemark[\ref{note3}],
%Libra R-CNN~\cite{pang2019libra}\footnote{https://github.com/OceanPang/Libra\_R-CNN}, 
%Guided Anchoring~\cite{wang2019region}~\footnotemark[\ref{note3}], 
TridentNet~\cite{li2019scale}, %\footnote{https://github.com/facebookresearch/detectron2}, 
SNIPER~\cite{singh2018sniper}\footnote{https://github.com/MahyarNajibi/SNIPER}, 
%Panoptic-FCN~\cite{kirillov2019panoptic}\footnotemark[\ref{note3}], 
R-FCN~\cite{dai2016r}\footnote{https://github.com/msracver/Deformable-ConvNets}, 
YOLOv3~\cite{redmon2018yolov3}, %\footnote{https://github.com/pjreddie/darknet}, 
RefineDet~\cite{zhang2018single}, %\footnote{https://github.com/sfzhang15/RefineDet} 
and SSD~\cite{liu2016ssd}\footnote{https://github.com/tensorflow/models/tree/master/research/object\_detection} having ResNet101~\cite{resnetHe15}, ResNext101~\cite{xie2017aggregated}, Triple-ResNeXt152, 
%\footnote{https://github.com/PKUbahuangliuhe/CBNet}, 
InceptionV2~\cite{normalization2015accelerating} or VGG16~\cite{Simonyan2015VeryRecognition} backbone-networks as our baseline benchmark algorithms on the test set for their excellent performance in object detection on ground images by\glspl{gls:HBB}. Furthermore, we modify the original Cascade Mask-RCNN to detect objects with\glspl{gls:RBB} described by $\{(x_i, y_i),~i= 1,2,3,4\}$. We further adapt the algorithm to able to detect objects with OBBs denoted as $(x_c, y_c, w, h, \theta)$, as $\theta$ means the vehicle head angle.
In order to evaluate the benchmark algorithms on EAGLE, we propose three different tasks including detection by HBB, RBB, and OBB.%\textbf{H}orizontal \textbf{B}ounding \textbf{B}oxes (HBB), \textbf{R}otated \textbf{B}ounding \textbf{B}oxes (RBB), and \textbf{O}riented \textbf{B}ounding \textbf{B}oxes (OBB).
As the evaluation metric, we employ~\gls{gls:map} similar to PASCAL VOC.
The image patches are stitched to form the original image before the evaluation step. In order to remove the redundant detected boxes in the overlapping regions as well as the patches themselves, we apply~\gls{gls:NMS} with a threshold of $0.3$ for HBB and $0.1$ for both RBB and OBB.
\subsection{Image splitting}
In the training phase, due to the large size of the images (5616$\times$3744~px) in the EAGLE dataset which cannot be fitted into the object detectors for the training process, we crop them into $1024\times1024$~px patches with a 50\% overlap in a sliding window fashion, resulting in 70 patches per image leading to 12075, 4025, and 8050 patches of training, validation and test respectively. The overlaps of the patches allows keeping all the objects, even if partially clipped at image boundaries. Patches thus ending up partially outside the image are shifted back into the image window. 
Patch-wise predictions are stitched into full images and overlaps were merged using NMS. 
This process could cut some vehicles into two parts. In this case, we compute the ratio between the area covered by each part ($A_i,~i= 1,2$) and that of the complete vehicle ($A_O$) as $U_i=A_i/A_O$ similar to~\cite{Xia2017DOTA}, but with the difference that we adapt the parts' ground truths to the image boundaries to have the highest intersection with the original object. After that, for $U_i\geqslant0.7$, the attribute of the part remains unchanged, for $0.1\le U_i < 0.7$, the attribute of the part is changed to "difficult", and for $U_i<0.1$, the part is ignored. %This implementation will be made public.
Moreover, the part which does not include the front part of the vehicle (depicting the orientation) is assigned a "difficult" flag to its orientation attribute.
For the testing step, we crop the images, but with a stride of $912$~px (10\% overlap  %with previous patch 
to ensure the coverage of the vehicles in their full appearance as well.
\subsection{\textbf{H}orizontal \textbf{B}ounding \textbf{B}oxes (HBB) baselines}
We generate the ground truth for\gls{gls:HBB} by calculating the center coordinates of the minimum and maximum in $x$ and $y$ coordinates in the original rotated bounding box ground truth. We train the baseline algorithms with their default settings and hyper-parameters for a fair comparison. Table~\ref{tab:ResultsHbbandRBB} shows the\gls{gls:HBB} detection results which indicates how challenging this dataset is for the-state-of the-art methods, with Cascade Mask-RCNN achieving the best performance of 39.29\% mAP. SSD and Yolov3 have very low performance compared to the others. This could be due to the random crops during data augmentation suggested by~\cite{Xia2017DOTA}. Furthermore, the results depict a considerable difference between the ground-level and aerial objects concerning their size, scale and appearance.
% \ResultsHBB
\subsection{\textbf{R}otated \textbf{B}ounding \textbf{B}oxes (RBB) baselines}
Since most of the state-of-the-art algorithms are designed for non-oriented objects, direct application of the algorithms for detecting the oriented-objects is not efficient which makes the benchmark of the existing algorithms for\gls{gls:RBB} challenging.
%As an approach to deal with this challenge, from the existing baseline algorithms, 
We select and modify the Cascade Mask-RCNN~\cite{cai2019cascade} algorithm for predicting rotated bounding boxes, due to its accuracy %on the MSCOCO dataset and particularly on small objects as well as 
on the\gls{gls:HBB} task of the EAGLE dataset. For the rest of algorithms, we train the algorithms on the\gls{gls:HBB} annotations of our dataset and test them on the\gls{gls:RBB} annotations.%, and compare their results.
Cascade Mask-RCNN is composed of one\gls{gls:RPN} and three detection and segmentation heads with thresholds $U=\{0.5,~0.6,~0.7\}$. While\gls{gls:RBB} ground truth is defined by $\{(v_{xi}, v_{yi}), i= 1,2,3,4\}$ vertices,\gls{gls:RPN} generates horizontal rectangles denoted by their top-left (TL) and bottom-right (BR) vertices $RoI=(x_{TL}, y_{TL}, x_{BR}, y_{BR})$. Therefore, we adapt the ground truth to rectangles by $x_{TL}=v_{x1} = v_{x4}$, $x_{BR} = v_{x2} = v_{x3}$, $y_{TL}=v_{y1} = v_{y4}$, and $y_{BR} = v_{y2} = v_{y3}$, similar to~\cite{Xia2017DOTA}. An alternative would be using rotated\gls{gls:RPN} as mentioned in~\cite{azimi2018towards}. However, we try to preserve the structure of the algorithm as much as possible.
In the detection heads, the output target $T=\{(t_xi, t_{yi}), i= 1,2,3,4\}$ for each RoI and its ground truth $G=\{(g_{xi}, g_{yi}), i= 1,2,3,4\}$ are defined as: 
\begin{align}
	t_{xi}   &= (g_{xi} - v_{xi})/w, &   t_{yi}   &= (g_{yi} - v_{yi})/h
\end{align}
where $w = x_{BR} - x_{TL}$ and $h = y_{BR} - y_{TL}$, similar to~\cite{liao2018textboxes++}. We consider the coordinates of each ground truth $G$ as the object mask to prepare the mask for the segmentation head. Table~\ref{tab:ResultsHbbandRBB} shows the results of the modified Cascade Mask-RCNN trained and tested on\gls{gls:RBB} compared with other baselines trained on\gls{gls:HBB} and tested based on\gls{gls:RBB} ground truth. We denote the modified method as \textit{Cascade Mask-RCNN-Rotated}. The results show that by adapting the algorithm to rotated bounding box detection, we can achieve an improvement of about 7\% mAP points. It also indicates that RBB task is a more difficult task than general HBB. 
  \begin{figure}
    \vspace{0.1cm}
 	\centering
 	\includegraphics[width=\columnwidth]{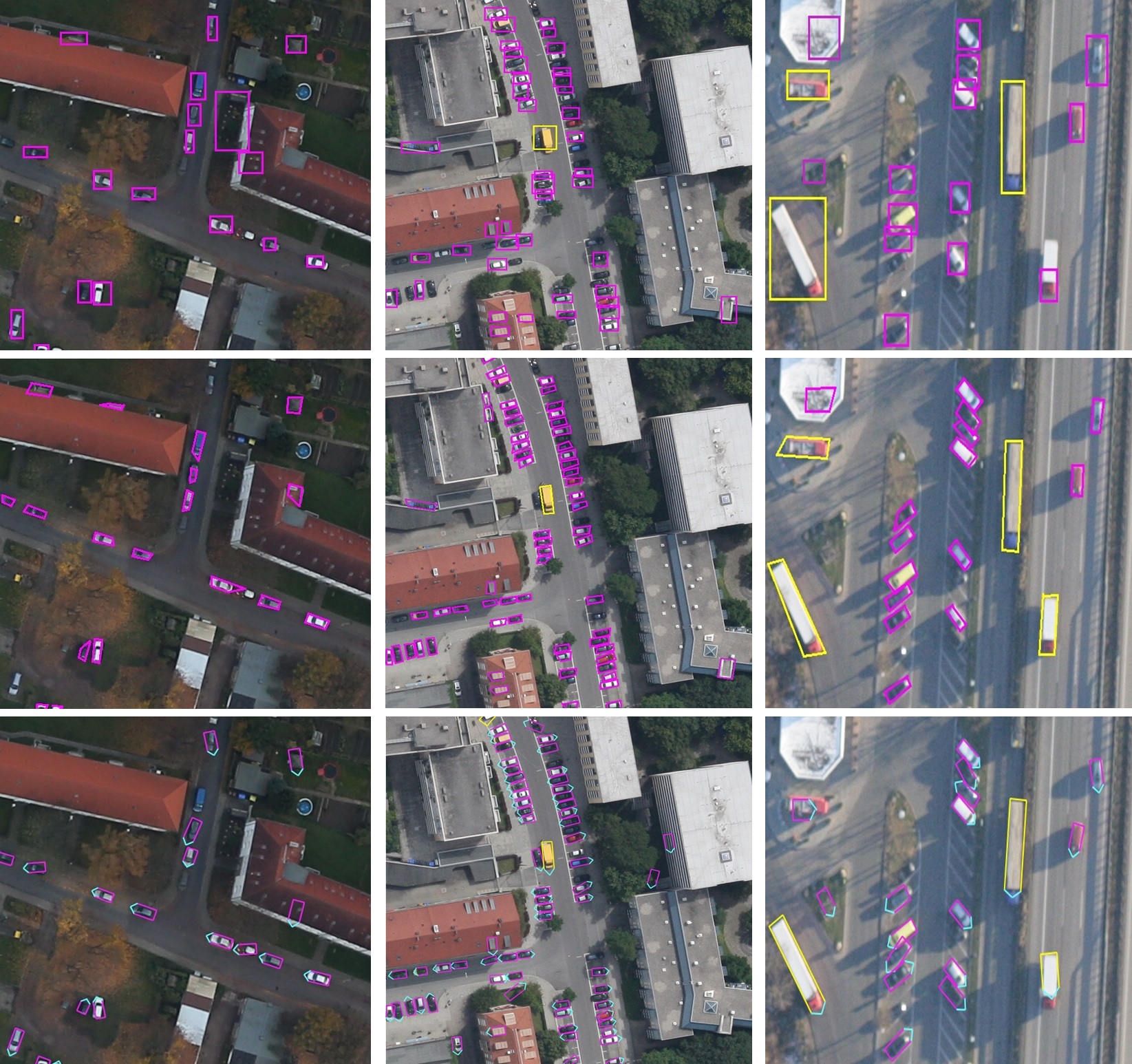}
 	\caption{Test prediction samples of Cascade Mask-RCNN trained on the EAGLE dataset. The first row is the result of horizontal bounding box (HBB), the middle row for rotated quadrilateral bounding boxes (RBB), and the bottom row is the result of oriented bounding boxes (OBB). Magenta is for small-vehicle and yellow for large-vehicle. The orientation is depicted in cyan.}
 	\label{fig:results}
 	\vspace{-0.2cm}
 \end{figure}
\ResultsHBBandRBB
% \ResultsRBB

\subsection{\textbf{O}riented \textbf{B}ounding \textbf{B}oxes (OBB) baselines}
For the benchmark based on\gls{gls:OBB}, we modify the detection heads of Cascade Mask-RCNN to predict the bounding box angles, and denote it as \textit{Cascade Mask-RCNN-Oriented}. To this end, We regress over $T=(x_c, y_c, w, h, \theta)$ instead of $(x_{TL}, y_{TL}, x_{BR}, y_{BR})$. Other possibilities are regression over $T=\{(x_i, y_i),~i= 1,2,3,4;~\theta)\}$ , or considering the clockwise order of bounding box vertices. The angle regression is defined as:
\begin{align}
	t_{\theta} = tan(g_\theta - v_\theta),
\end{align}
where $tangent$ function is used to ensure the periodicity of the angle regression, but other regression approaches can be considered. Similar to the Fast-RCNN~\cite{fastrcnn} algorithm, we use the smooth $L_1$ loss for bounding box regression and Cross-entropy loss for classification.
We evaluate the performance of the algorithm on the\gls{gls:OBB} task by comparing the center coordinates, angle, width and height of predicted oriented bounding box. For orientation estimation, we divide the angles in the range of $(-180, 180)$ into 16 output bins and we consider an angle prediction to be correct if it falls into the same bin as the ground truth. Cascade Mask-RCNN-Oriented achieves 43.87\%\gls{gls:map} which is 59.45\%\gls{gls:ap} and 28.29\%\gls{gls:ap} for small and large vehicle and with the angle accuracy of 67.34\%.

\subsection{Experimental analysis}
By analyzing the results shown in Table~\ref{tab:ResultsHbbandRBB}, we observe that the HBB detection is still challenging with respect to very small size objects, densely crowded regions, and occlusions in aerial images. In Figure~\ref{fig:results}, we provide a comparison of small and large vehicle detection methods of\gls{gls:HBB},\gls{gls:RBB}, and\gls{gls:OBB}.
As shown in Figure~\ref{fig:results}, for areas in which vehicles are parked tightly  %closed to each other
, we observe that\gls{gls:HBB} is less accurate than\gls{gls:RBB} and\gls{gls:OBB} in precise localization of vehicles in which several detection results are suppressed by\gls{gls:NMS} and other post-processing steps. Furthermore, we see that 
%in these areas that 
some vehicles do not have right-angle detections for the RBB task leading to mistakes in the localization while\gls{gls:OBB} does not have this issue, resulting in a better performance. Therefore\gls{gls:OBB} is the more accurate way in oriented object detection in aerial images.% such as vehicles in real-world scenarios. 
As for false positives, some non-vehicles objects appear similar to vehicles, confusing detectors as shown in the left column of Figure~\ref{fig:results}, showing false positives over the roofs. Also in the results of RBB in the middle column, a trash bin was detected as small vehicle. The less accuracy of the detector in large-vehicle detection compared to small-vehicle is the higher number of small-vehicle instances compared to large-vehicle ones leading to an unbalanced dataset. %Hard Negative Mining~\cite{AbhinavShrivastava2016} could resolve this issue considerably. 
Also, in highly dense areas, results of both\gls{gls:RBB} and\gls{gls:OBB} are not satisfying implying the high difficulty of this task.% for the-state-of-the-art detectors.
\Resultssecondsplit

\subsection{Impact of data-related factors on the performance} The smaller\gls{gls:GSD} is already known to improve performance drastically~\cite{shermeyer2019effects,azimi2018towards}, but requires very-high resolution image acquisition, which may not always be possible. Smaller size and scale can also degrade the performance. The segmentation of objects down to 2px-wide at different scales was already successfully presented~\cite{azimi2019skyscapes}. Experiments on EAGLE indicates other challenges such as low-illumination, haze, shadow and occlusion as critical factors preventing state-of-the-art object detectors from performing well. EAGLE will support future works aiming at solving these real-world issues.
\subsection{Cross-dataset validation}
We do a cross-dataset generalization to evaluate the generalization capability of EAGLE dataset. We select DOTA for comparison and its validation set for testing. 
%as the dataset with the highest number of vehicle instances in aerial images after~\gls{EAGLE} and we test on DOTA validation set for the comparison as the ground truth for the test set is not available. 
We choose Cascade Mask-RCNN for validation experiments with~\gls{gls:HBB} ground truth. 
\Resultscrossvalidation
Table~\ref{tab:Resultscrossvalidation} shows that a model trained on EAGLE generalizes well to DOTA, scoring only 6\% mAP below a model trained on DOTA, indicating that~\gls{EAGLE} contains features of DOTA to a large extend. Moreover, as the annotation quality in EAGLE is significantly higher than in DOTA specially with respect to very small vehicles (as mentioned in Section~\ref{sec:imageannotation}), a portion of false positives in this comparison is due to the detection of vehicles which are generally not annotated and ignored in DOTA, due to their small size.
As for DOTA, the model trained on it only achieves 28.23\% mAP on EAGLE (-11\% mAP of the model trained on EAGLE) reflecting that EAGLE is significantly more diverse and challenging than the current available datasets which makes it appropriate for real-world vehicle detection scenarios.
\section{Conclusion}
We present\gls{EAGLE}, a large-scale dataset for task of vehicle detection in aerial imagery, which is multiple times larger than existing datasets. Unlike common object detection datasets, we provide a high number of annotated instances with oriented bounding boxes. We build a dataset specifically focusing on real-world scenarios which includes a variety of situations in aerial photography such as time, weather, and places. The detection of vehicles in any situation regardless of their size and appearance with arbitrary orientations contains useful information for different applications%fast situation awareness for instance in disaster scenarios. We presume that this data will be
, making it useful for many practical applications. %Although this dataset contains usual aerial scenes, 
Our benchmarks show%it remains a very 
\gls{EAGLE} is a very challenging dataset for the current state-of-the-art object detection algorithms. We also showcase a general method on object detection which can be modified to detect oriented objects. We believe\gls{EAGLE} addresses the task of vehicle detection in remote vision bringing it to the next practical level. It also introduces interesting challenges to object detection domain in computer vision.

\section{Acknowledgement}
We thank Ternow AI GmbH for their kind support. 
%The authors would like to thank the National Center for Airborne Laser Mapping and the Hyperspectral Image Analysis Laboratory at the University of Houston for acquiring and providing the data used in this study, and the IEEE GRSS Image Analysis and Data Fusion Technical Committee.
\bibliographystyle{IEEEtran}
\bibliography{bibliography.bib}
\end{document}